\documentclass[letterpaper, 10 pt, conference]{ieeeconf}  

\IEEEoverridecommandlockouts                              

\usepackage[export]{adjustbox}
\usepackage{graphicx}
\usepackage{amsmath,amssymb,amsfonts}

\usepackage{xcolor} 

\renewcommand{\textcolor}[2]{#2}
\usepackage{booktabs}
\usepackage{tabularx}
\usepackage{array}
\usepackage{multirow}
\usepackage{makecell}

\usepackage{siunitx}
\newcommand{\best}[1]{\textbf{#1}}
\newcommand{\na}{\textemdash}

\usepackage{algorithm}
\usepackage{algpseudocode}

\usepackage{placeins} 

\usepackage{cite} 
\usepackage[
  colorlinks=true,
  linkcolor=black,
  citecolor=black,
  urlcolor=black
]{hyperref}

\title{\LARGE \bf
\textcolor{blue}{DAGS-SLAM: Dynamic-Aware 3DGS SLAM via Spatiotemporal Motion Probability and Uncertainty-Aware Scheduling}
\thanks{This work was supported by the National Natural Science Foundation of China (61972131, 62332016).}
}
\author{~Li~Zhang,~Yu-An~Liu,~Xijia~Jiang,~Conghao~Huang,~Danyang~Li,~Yanyong~Zhang,~\IEEEmembership{Fellow,~IEEE}
\thanks{L. Zhang, Y. Liu, X. Jiang, and C. Huang are with the School of Mathematics, Hefei University of Technology, Hefei 230601, China (e-mail: lizhang@hfut.edu.cn; yuanliu0015@gmail.com; xijiajiang@mail.hfut.edu.cn; 2023111354@mail.hfut.edu.cn.
\newline \hspace*{2em}D. Li is with the School of Software, Tsinghua University, Beijing 100084, China (e-mail: lidanyang1919@gmail.com).
\newline \hspace*{2em}Y. Zhang is with  the School of Computer Science and Technology, University of Science and Technology of China, Hefei 230026, China (e-mail: yanyongz@ustc.edu.cn).}}

\begin{document}

\raggedbottom
\flushbottom
\maketitle
\thispagestyle{empty}
\pagestyle{empty}

\begin{abstract}
\textcolor{blue}{Mobile robots and AR/IoT devices demand real-time localization and dense reconstruction under tight compute and energy budgets. While 3D Gaussian Splatting (3DGS) enables efficient dense SLAM, dynamic objects and occlusions still degrade tracking and mapping. Existing dynamic 3DGS-SLAM often relies on heavy optical flow and/or per-frame segmentation, which is costly for mobile/edge deployment and brittle under challenging illumination. We present DAGS-SLAM, a dynamic-aware 3DGS-SLAM system that maintains a spatiotemporal motion probability (MP) state per Gaussian and triggers semantics on demand via an uncertainty-aware scheduler.} DAGS-SLAM fuses lightweight YOLO instance priors with geometric cues to \textcolor{blue}{estimate and temporally update MP}, propagates MP to the front-end for dynamic-aware correspondence selection, and suppresses dynamic artifacts in the back-end via MP-guided optimization. Experiments on public dynamic RGB-D benchmarks show improved reconstruction and robust tracking while sustaining real-time throughput on a commodity GPU, \textcolor{blue}{demonstrating a practical speed–accuracy trade-off with reduced semantic invocations toward mobile/edge deployment.}
\end{abstract}

\section{INTRODUCTION}

\textcolor{blue}{Spatial understanding is a core capability for mobile agents (e.g., robots, drones, and AR devices) operating in unknown environments, underpinning localization, mapping, and task-oriented navigation~\cite{zheng2025graphgeo, li2025analyzing}. Early studies on location awareness demonstrated that lightweight spatial cues can help organize computation and improve efficiency in distributed settings~\cite{liu2004location, liu2005location}. Modern embodied agents, however, further require geometrically precise localization and dense scene modeling under strict compute budgets.}
Simultaneous Localization and Mapping (SLAM) provides this capability by jointly estimating the agent’s pose and building a representation of the surrounding scene~\cite{davison2007monoslam}. Meanwhile, recent advances in 3D Gaussian Splatting (3DGS) have demonstrated high-fidelity appearance modeling and efficient rendering for novel view synthesis, motivating a new line of dense SLAM systems built on 3DGS~\cite{10657581, deng2024compact}. However, most existing 3DGS-SLAM pipelines implicitly assume static scenes. In real-world mobile scenarios with moving objects and frequent occlusions, this assumption can lead to unstable tracking and dynamic artifacts in reconstructed maps~\cite{zhou2024feature, lu2024scaffold}.

\begin{figure}[t]
  \centering
  \setlength{\tabcolsep}{2pt}
  \renewcommand{\arraystretch}{0}
  \begin{tabular}{cc}
    \includegraphics[width=0.49\linewidth]{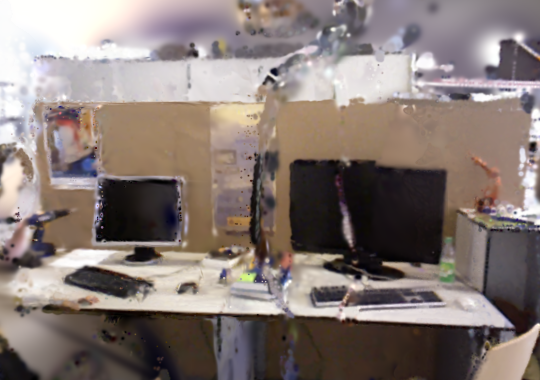} &
    \includegraphics[width=0.49\linewidth]{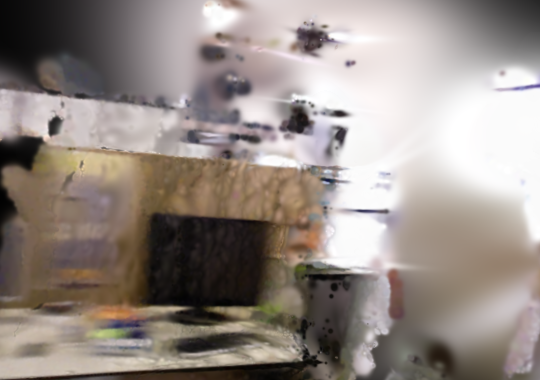} \\
    \includegraphics[width=0.49\linewidth]{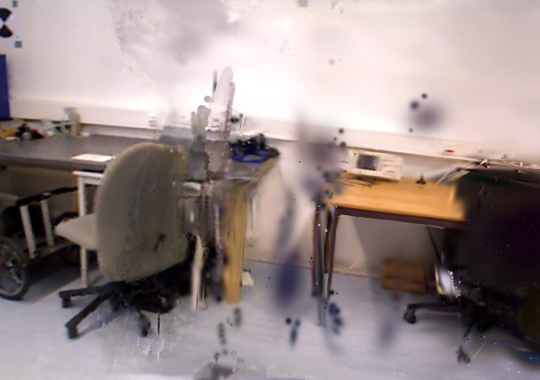} &
    \includegraphics[width=0.49\linewidth]{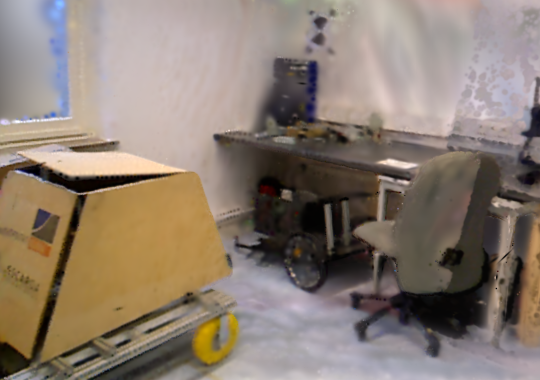} \\
  \end{tabular}
  \vspace{-2pt}
  \caption{\textcolor{blue}{Typical failure cases of 3DGS-SLAM in dynamic scenes. Moving objects are often fused into the Gaussian map, leading to ghosting/blobs, smeared textures, and broken surfaces, which in turn degrades both tracking stability and static-scene reconstruction quality.}}
  \label{fig:gs_dynamic_fail}
  \vspace{-6pt}
\end{figure}

Dynamic scenes challenge 3DGS-SLAM for three reasons: (i) moving objects introduce inconsistent multi-view constraints, causing the map to overfit transient motion; (ii) dynamic regions can dominate pixel-level rendering residuals and bias pose updates; and (iii) under occlusion and blur, motion boundaries become ambiguous, so hard dynamic pruning may mistakenly remove static structure near boundaries. Consequently, dynamic scenes violate the static-world assumption of most 3DGS-SLAM systems: the same 3D location can receive contradictory appearance and depth observations over time. As shown in Fig.~\ref{fig:gs_dynamic_fail}, these artifacts degrade reconstruction and view synthesis and further corrupt tracking cues, resulting in drift bursts or instability. This motivates dynamic-aware Gaussian mapping and robust pose estimation that suppress dynamic primitives while preserving static structure.
\textcolor{blue}{Importantly, for edge-motivated SLAM, achieving such robustness also calls for avoiding heavyweight modules under tight latency and energy budgets.}

Recent methods address dynamics by leveraging learned motion cues, typically using deep optical flow to produce per-frame dynamic masks~\cite{kong2024dgs}. These masks are then used to gate tracking and rendering, often together with additional rendering losses to suppress dynamic content. While effective in some settings, this design raises two concerns for resource-constrained SLAM. First, optical-flow-based modules are computationally intensive~\cite{zhang2020flowfusion}, which can make sustained real-time execution under tight compute budgets difficult and motivates system-level offloading and scheduling for low-latency SLAM~\cite{zhang2025orchestrating}. Second, end-to-end performance depends heavily on mask quality: under illumination changes, fast motion, and occlusions, motion masks can become noisy or incomplete~\cite{shi2026codeocr, chen2025progressive}. Such errors may propagate into correspondence selection and rendering optimization, ultimately degrading tracking stability. As a result, these pipelines can be sensitive on challenging sequences and may yield degraded tracking accuracy under adverse conditions~\cite{cheng2022sg}.

\textcolor{blue}{Our goal is to build a dynamic-robust 3DGS-SLAM system that is real-time and resource-efficient, motivated by mobile/edge constraints. Such constraints are increasingly emphasized in edge-assisted SLAM, where on-chip intelligence and system co-design are used to reduce end-to-end latency and energy consumption on mobile platforms~\cite{li2024reshaping}. In a similar spirit, prior studies on lightweight sensing and computation highlight a broader trend toward practical, resource-aware localization~\cite{wei2021imag+}. Motivated by these constraints, we make the following observation: rather than making a hard dynamic/static decision at every frame, it can be more stable to maintain a temporally updated soft motion state. This state smooths noisy and occasionally stale observations while remaining responsive to motion changes, and it provides a unified interface to influence both front-end tracking decisions and back-end rendering/optimization.}

Based on this observation, we propose DAGS-SLAM, an efficient dynamic-robust 3DGS-SLAM system with an edge-motivated, resource-aware design. Each Gaussian is augmented with a motion probability (MP) attribute~\cite{wu2024mpoc},
\textcolor{blue}{maintained as a temporally updated soft state to reduce brittle decisions under occlusions and imperfect priors, instead of aggressively discarding dynamic evidence in the front-end~\cite{li2025dff}.}
We estimate MP by fusing lightweight instance priors~\cite{wang2023yolov7,abdelnasser2015semanticslam} with geometric cues, and propagate MP to tracking for dynamic-aware correspondence selection.
\textcolor{blue}{To reduce semantic overhead, we introduce an uncertainty-aware semantic-on-demand scheduler that avoids unnecessary instance-prior invocations when the motion state is confident.}
We additionally recover static structure near motion boundaries via epipolar verification and suppress dynamic artifacts through MP-guided rendering/optimization.

Our main contributions are summarized as follows:
\begin{itemize}
  \item We propose DAGS-SLAM, a dynamic-robust 3DGS-SLAM framework that augments each Gaussian with a temporally updated MP state and propagates MP to tracking for dynamic-aware correspondence selection.
  \item \textcolor{blue}{We develop an MP estimation and resource-aware semantic design that fuses instance-level semantic priors with geometric cues, and uses an uncertainty-aware semantic-on-demand scheduler to reduce semantic runtime cost under mobile/edge-motivated constraints.}
  \item We enhance robustness and reconstruction quality by combining (i) epipolar-geometry verification with static recovery and densification near motion boundaries/occlusions, and (ii) MP-guided rendering optimization with photometric/depth reweighting, MP regularization, and an edge-warp loss to suppress dynamic artifacts under noisy priors.
\end{itemize}

\section{RELATED WORK}

\begin{figure*}[!t]
\centering
\includegraphics[width=\textwidth,height=0.78\textheight,keepaspectratio]{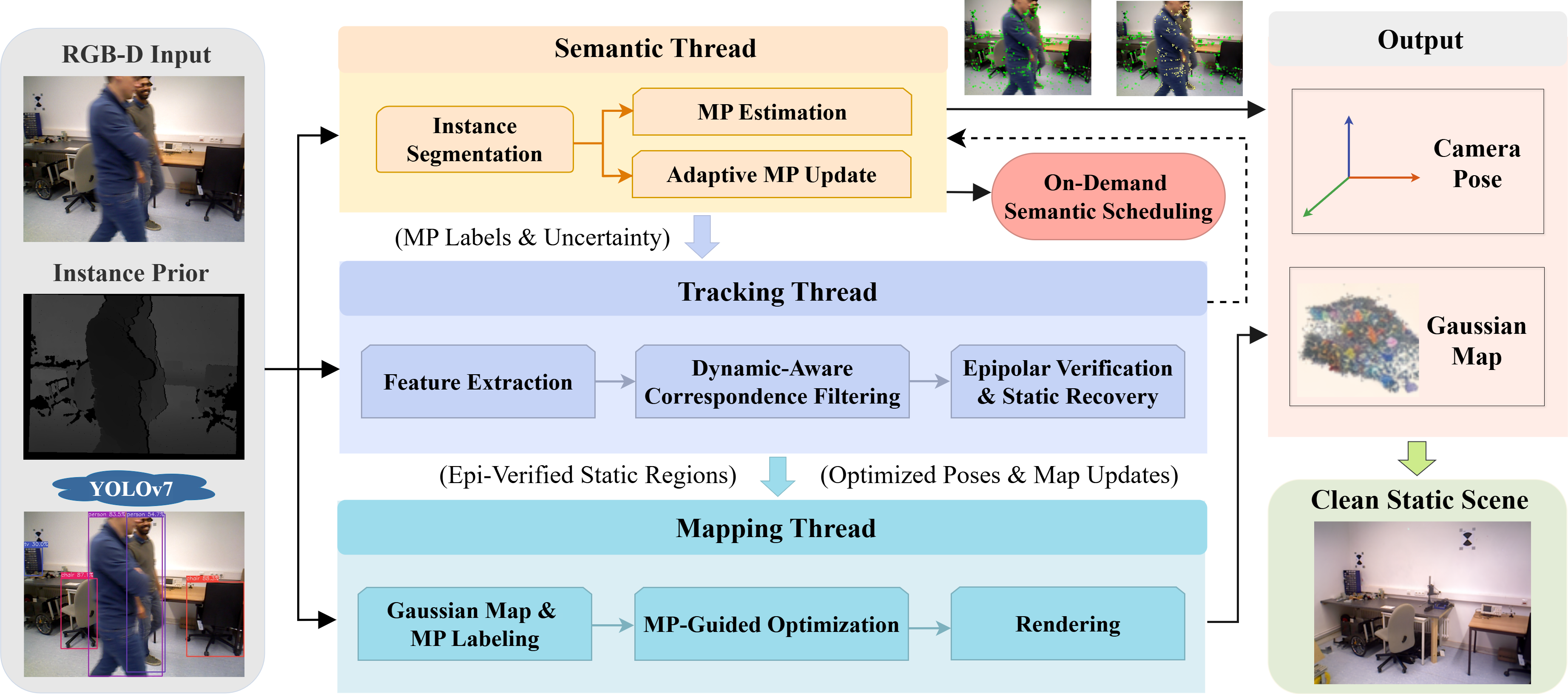}
\caption{System pipeline of DAGS-SLAM. A semantic thread maintains a temporally updated per-Gaussian MP state by fusing lightweight instance priors with geometric cues, and \textcolor{blue}{an uncertainty-aware scheduler triggers semantics on demand to reduce overhead}. MP is propagated to the tracking thread for dynamic-aware correspondence filtering and to the mapping thread for MP-guided rendering optimization, producing clean static-scene reconstructions under dynamic interference.}
\label{fig:twocolumn}
\end{figure*}

\subsection{NeRF-based SLAM in dynamic environments}

Recent progress in Neural Radiance Fields (NeRF)~\cite{mildenhall2021nerf} has inspired NeRF-based SLAM systems for online reconstruction and tracking~\cite{li2023end,wang2021nerf}. iMAP~\cite{sucar2021imap} represents a scene with a single MLP in an online RGB-D SLAM pipeline, while NeRF-SLAM improves monocular reconstruction quality via hierarchical representations and uncertainty-aware losses~\cite{ortiz2022isdf}. Despite promising results in controlled settings, NeRF-style implicit representations can be challenging in real dynamic scenes and resource-limited deployments, due to the coupled difficulties of modeling non-static observations and maintaining real-time performance.
To address dynamics, NID-SLAM~\cite{xu2024nid} leverages optical-flow-based motion estimation to generate dynamic masks and performs background completion, aiming to reduce dynamic interference. RoDyn-SLAM~\cite{jiang2024rodyn} combines motion and semantic cues to refine motion estimation, while DDN-SLAM~\cite{li2024ddn} integrates semantic features with probabilistic modeling and sampling strategies to mitigate dynamic effects. Nonetheless, these systems may still suffer from either imperfect motion/mask cues or limited real-time throughput, and residual inconsistencies caused by dynamics can remain difficult to fully eliminate in practice.

\subsection{Dynamic SLAM based on 3DGS}

3DGS provides an explicit point-based representation with efficient rendering and high-fidelity appearance modeling~\cite{li2024sgs, zhou2024mod, li2025densesplat}. Compared to NeRF-style implicit MLPs, 3DGS often enables faster rendering and more direct control over geometric primitives, making it attractive for dense SLAM. Early 3DGS-SLAM systems, such as SplaTAM~\cite{keetha2024splatam} and MonoGS~\cite{matsuki2024gaussian}, demonstrate real-time reconstruction and tracking with geometric verification. However, most representative 3DGS-SLAM pipelines still assume static scenes, and moving objects can induce pose drift and persistent dynamic artifacts when they dominate observations.
Dynamic 3DGS-SLAM remains relatively under-explored. DG-SLAM~\cite{xu2024dg} introduces motion-aware masking and dynamic Gaussian handling to improve tracking and rendering quality for static regions, yet its performance can be sensitive to the precision of semantic/motion priors, especially near motion boundaries and under occlusion. More broadly, dynamic 3DGS-SLAM methods need to address two coupled issues: (i) incorrect dynamic/static decisions may remove static structures near boundaries and create holes in the map; and (ii) tracking and mapping objectives can become inconsistent when front-end feature filtering and back-end rendering-based optimization respond to dynamics differently, which complicates stable optimization under resource constraints.

\section{SYSTEM OVERVIEW}
Fig.~\ref{fig:twocolumn} overviews DAGS-SLAM, a three-thread pipeline designed for dynamic-robust 3DGS-SLAM under mobile/edge resource budgets. Given an RGB-D stream, the Semantic Thread estimates and temporally updates per-Gaussian MP by fusing lightweight instance priors and geometric cues. \textcolor{blue}{To reduce runtime overhead, an uncertainty-aware semantic-on-demand scheduler refreshes the instance prior only when MP uncertainty and/or tracking--geometry inconsistency indicates potential dynamic disturbance.} The Tracking Thread associates MP-labeled Gaussians with image features to perform dynamic-aware correspondence filtering and preserve reliable static matches for pose estimation. The Mapping Thread jointly optimizes camera poses and Gaussian parameters with MP-guided rendering objectives to suppress dynamic artifacts and produce clean static-scene reconstructions. Details of MP estimation, verification, and loss design are presented in subsequent sections.

\section{METHOD}\label{sec:method}
\subsection{Dynamic Gaussian Segmentation Based on MP}\label{sec:mp_seg}
\vspace{0.3\baselineskip}
\begin{figure}[t]
\centering
\includegraphics[width=\columnwidth]{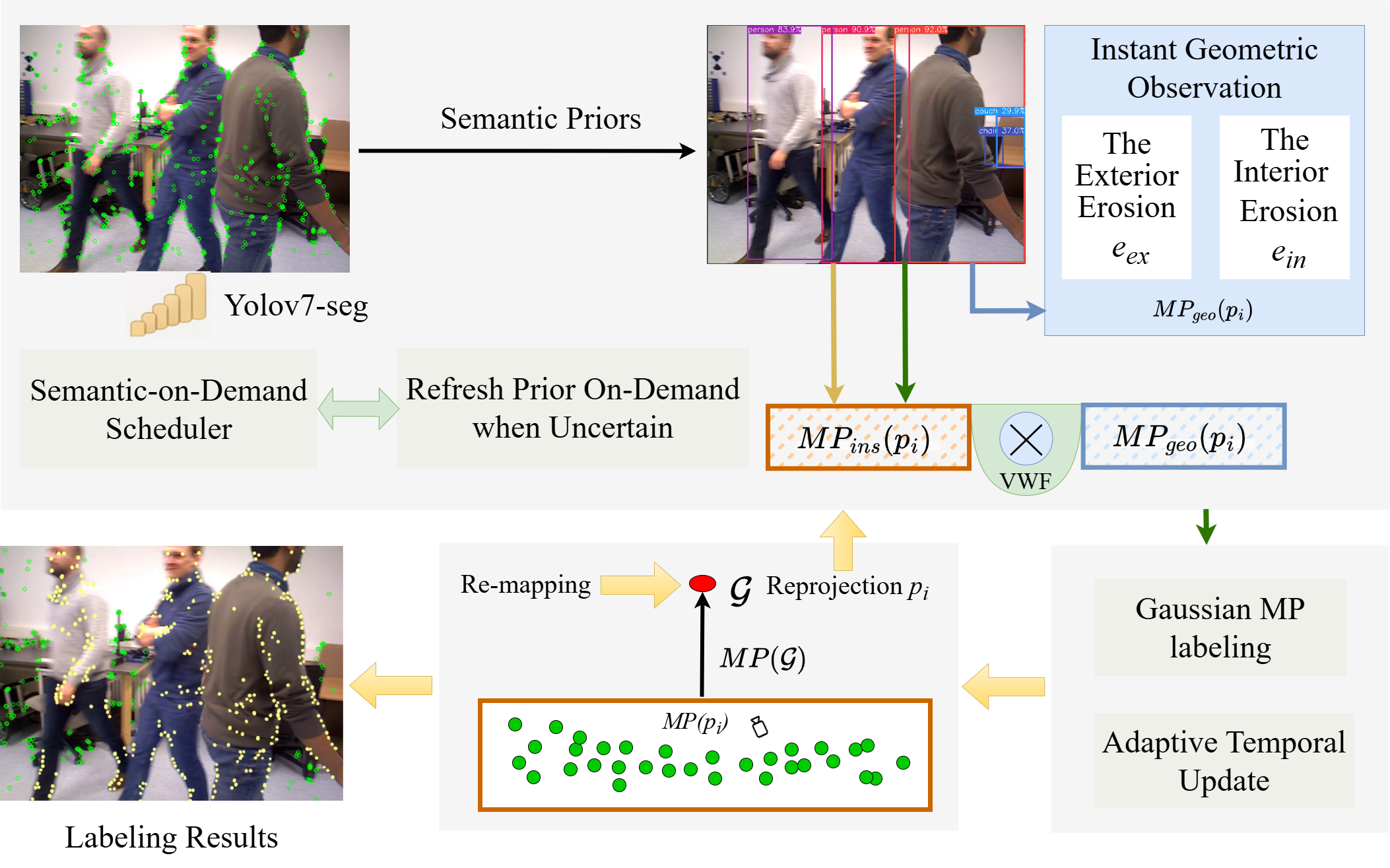}
\vspace{-2pt}
\caption{Illustration of the proposed labeling process, where yellow denotes dynamic outliers and green denotes static features.}
\label{fig:labeling}
\vspace{-4pt}
\end{figure}

As illustrated in Fig.~\ref{fig:labeling}, this module maintains a temporally updated per-Gaussian MP state that supports dynamic-aware labeling in the back-end and propagates to the front-end for correspondence filtering during tracking.
\textcolor{blue}{By treating motion as a soft state rather than per-frame hard decisions, MP smooths noisy observations while remaining responsive to persistent motion changes, which is crucial under occlusion and imperfect priors.}

In 3DGS-SLAM, each Gaussian primitive $\mathcal{G}$ is parameterized by its 3D mean, opacity, scale, and rotation.
To cope with dynamic scenes while preserving accurate pose estimation and high-fidelity rendering, we augment each Gaussian with the MP state.
Rather than filtering dynamic evidence only in the front-end, we infer dynamic/static labels in the back-end using per-Gaussian MP, suppress dynamic Gaussians during optimization and rendering, and propagate the Gaussian-level MP to guide correspondence selection in the front-end.

\paragraph{Gaussian--feature association.}
To connect the Gaussian-level MP state to feature tracking with minimal overhead, we associate each Gaussian with nearby tracked features through a multi-scale image pyramid $\{\mathcal{I}^{s}\}$.
For a Gaussian $\mathcal{G}$ whose mean projects to pixel $\pi(\mathcal{G})$ in the current frame, candidate features at each scale $s$ are collected within a radius $r_s$ around the projected location:
\begin{equation}
\mathcal{N}^{s}(\mathcal{G})=\{p \mid \|p-\pi^{s}(\mathcal{G})\|\le r_s\}.
\end{equation}
Let $\mathcal{N}(\mathcal{G})=\bigcup_s\mathcal{N}^s(\mathcal{G})$.
We further assign a soft association weight $w(p,\mathcal{G})$ using a multi-scale Gaussian kernel, so that closer features contribute more to the Gaussian-level aggregation~\cite{huang2024photo}.

\paragraph{Instantaneous MP observation from semantic and geometric cues.}
At time $t$, an instantaneous MP observation is estimated by combining (i) an instance-based semantic prior and (ii) keyframe-based geometric inconsistency cues, so that the system remains robust when either cue becomes unreliable.

Given an instance $k$ with mask region $\Omega_k$, category $\mathrm{cls}_k$, and confidence $c_k$, we define a semantic motion prior for each feature point $p$ as
\begin{equation}
MP_{\mathrm{ins}}(p)=
\begin{cases}
\mu(\mathrm{cls}_k), & p \in \Omega_k,\\
\mu(\mathrm{bg}), & \text{otherwise},
\end{cases}
\end{equation}
where $\mu(\cdot)\in[0,1]$ is a class-to-motion prior table and ``bg'' denotes the background.
The semantic confidence is set as $\Gamma_{p}^{\mathrm{ins}}=c_k$ if $p\in\Omega_k$ and a small constant $c_{\mathrm{bg}}$ otherwise.

For geometric cues, we select a co-visible historical keyframe $t^\star$ and compute the reprojection error of $p$ with respect to its associated Gaussian $\mathcal{G}$:
\begin{equation}
e(p;\mathbf{R},\mathbf{t})=\left\|p-\pi\!\left(\mathbf{R}\,\mu_{\mathcal{G}}+\mathbf{t}\right)\right\|,
\label{eq:reproj_err}
\end{equation}
where $\mu_{\mathcal{G}}\in\mathbb{R}^3$ is the mean of $\mathcal{G}$ and $(\mathbf{R},\mathbf{t})$ is the relative pose.
Within an instance region $\Omega_k$, we compute robust statistics
\begin{equation}
\begin{aligned}
\mu_e &= \mathrm{Mean}\big(\{e(p)\}_{p\in\Omega_k}\big),\\
\sigma_e &= \mathrm{Std}\big(\{e(p)\}_{p\in\Omega_k}\big),\\
e_{\text{ub}} &= \mu_e + \lambda \sigma_e,
\end{aligned}
\label{eq:e_stats}
\end{equation}
and define a normalized geometric motion probability
\begin{equation}
MP_{\mathrm{geo}}(p)=\mathrm{clip}\!\left(\frac{e(p)}{e_{\text{ub}}},\,0,\,1\right).
\label{eq:mp_geo}
\end{equation}
When $\sigma_e$ is small, we use the instance-mean reprojection error to reduce boundary-induced fluctuations.

\paragraph{Fusion and Gaussian-level MP observation.}
We fuse semantic and geometric MPs at the feature level as
\begin{equation}
M_p = (1-K_p)\,MP_{\mathrm{ins}}(p)+K_p\,MP_{\mathrm{geo}}(p),
\label{eq:vwf}
\end{equation}
with the fusion gain
\begin{equation}
K_p=\frac{\Gamma_p^{\mathrm{geo}}}{\Gamma_p^{\mathrm{ins}}+\Gamma_p^{\mathrm{geo}}}\in[0,1],
\label{eq:gain}
\end{equation}
where $\Gamma_p^{\mathrm{geo}}$ reflects geometric reliability.
In practice, we set $\Gamma_p^{\mathrm{geo}}=\exp(-\sigma_{e,k}/\sigma_g)$ using the reprojection-error dispersion $\sigma_{e,k}$ within $\Omega_k$ (or within the associated neighborhood when no instance is available), so that more consistent geometry yields higher weight.
The instantaneous Gaussian-level MP observation is then obtained by aggregating associated feature MPs:
\begin{equation}
\hat{M}_\mathcal{G}^{t}
=\frac{\sum_{p\in\mathcal{N}(\mathcal{G})} w(p,\mathcal{G})\,M_p}
       {\sum_{p\in\mathcal{N}(\mathcal{G})} w(p,\mathcal{G})}.
\label{eq:mp_obs}
\end{equation}

\paragraph{\textcolor{blue}{Adaptive temporal MP update and dynamic labeling.}}
\textcolor{blue}{We maintain MP as a temporal state and update it via exponential smoothing:
\begin{equation}
M_\mathcal{G}^{t}=(1-\alpha_\mathcal{G}^{t})\,M_\mathcal{G}^{t-1}
+\alpha_\mathcal{G}^{t}\,\hat{M}_\mathcal{G}^{t}.
\label{eq:mp_temporal}
\end{equation}
To account for observation ambiguity and local inconsistency, we define an uncertainty term
\begin{equation}
\begin{aligned}
U_{\mathcal G}^{t}
&=
\underbrace{-M_{\mathcal G}^{t-1}\log(M_{\mathcal G}^{t-1}{+}\epsilon)
-(1{-}M_{\mathcal G}^{t-1})\log(1{-}M_{\mathcal G}^{t-1}{+}\epsilon)}_{\text{entropy}}\\
&\quad+\lambda_u\,\mathrm{Var}\!\left(\{M_p\}_{p\in\mathcal{N}(\mathcal{G})}\right),
\end{aligned}
\label{eq:mp_uncertainty}
\end{equation}
where $\epsilon$ avoids numerical issues and $\tilde U_\mathcal{G}^{t}$ is normalized to $[0,1]$ within the frame.
The update rate is set to
\begin{equation}
\begin{aligned}
\alpha_\mathcal{G}^{t}
&=\mathrm{clip}\!\Big(
\alpha_{\min}+(\alpha_{\max}-\alpha_{\min})\,(1-\tilde U_\mathcal{G}^{t})\,C_\mathcal{G}^{t},\\
&\qquad\qquad\ \alpha_{\min},\alpha_{\max}
\Big),
\end{aligned}
\label{eq:alpha_rule}
\end{equation}
where $C_\mathcal{G}^{t}=\exp\!\big(-\sigma_{e,\mathcal{G}}/\sigma_c\big)\in(0,1]$ measures geometric consistency and
$\sigma_{e,\mathcal{G}}=\mathrm{Std}\big(\{e(p)\}_{p\in\mathcal{N}(\mathcal{G})}\big)$.
This design increases $\alpha_\mathcal{G}^{t}$ when observations are reliable and geometry is consistent, and reduces it under high uncertainty.}

For Gaussians observed in the current frame, we compute the median MP $\tilde{M}$ and set an adaptive threshold
\begin{equation}
\tau = \max(\tilde{M},\,0.5).
\label{eq:tau}
\end{equation}
A Gaussian $\mathcal{G}_i$ is labeled as dynamic if $M_{\mathcal{G}_i}^{t}>\tau$; otherwise it is treated as static.

\subsection{\textcolor{blue}{Uncertainty-Aware Semantic-on-Demand Scheduling}}\label{sec:sod}
\begin{figure}[t]
\centering
\includegraphics[width=\columnwidth]{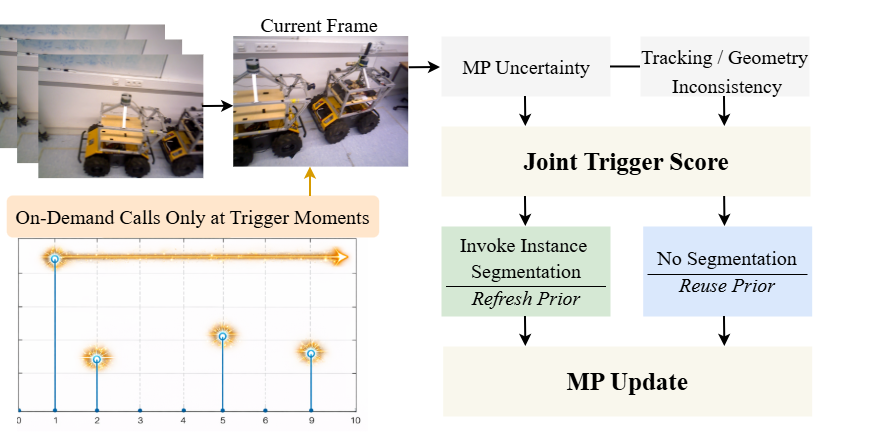}
\vspace{-2pt}
\caption{\textcolor{blue}{The uncertainty-aware semantic-on-demand scheduling mechanism.
MP uncertainty and tracking/geometry inconsistency are fused into a trigger score to decide whether to refresh the instance prior. Segmentation is invoked only at trigger moments; otherwise the prior is reused and MP is updated temporally.}}
\label{fig:tri}
\vspace{-4pt}
\end{figure}

\textcolor{blue}{Instance priors provide useful cues for MP estimation, yet dense segmentation at every frame/keyframe is often redundant and may become a major computational overhead.
Segmentation typically dominates the semantic runtime, making always-on invocation prohibitive on resource-limited mobile/edge devices.
As illustrated in Fig.~\ref{fig:tri}, we therefore adopt an uncertainty-aware semantic-on-demand strategy: the instance prior is refreshed only when the current estimates indicate reduced reliability; otherwise, the most recent prior is reused while MP is propagated via the temporal update.}

\paragraph{\textcolor{blue}{Trigger signals.}}
\textcolor{blue}{We consider two complementary indicators.}

\textcolor{blue}{\noindent(i) MP uncertainty.
We reuse the per-Gaussian uncertainty and summarize it at the frame level using a robust statistic:
\begin{equation}
\overline{U}^{t}=\mathrm{Median}\Big(\big\{\tilde U_{\mathcal{G}}^{t}\big\}_{\mathcal{G}\in\mathcal{V}^t}\Big)\in[0,1],
\label{eq:U_bar}
\end{equation}
where $\mathcal{V}^t$ denotes the set of Gaussians visible in frame $t$.
We use the median for robustness to outlier Gaussians caused by transient dynamics or occlusions.}

\textcolor{blue}{\noindent(ii) Tracking/geometry inconsistency.
From the tracking module, we compute a frame-level inconsistency score $R^{t}\in[0,1]$ based on the normalized mean reprojection residual:
\begin{equation}
R^{t}=\mathrm{clip}\!\left(\frac{\bar e^{t}}{e_{\mathrm{ref}}},\,0,\,1\right),
\label{eq:R_def}
\end{equation}
where $\bar e^{t}$ is the mean reprojection error over verified correspondences and $e_{\mathrm{ref}}$ is a reference constant.}

\paragraph{\textcolor{blue}{Joint trigger and scheduling rule.}}
\textcolor{blue}{The two signals are combined into a joint trigger score
\begin{equation}
S^{t}=w_u\,\overline{U}^{t}+w_r\,R^{t},
\label{eq:trigger_score}
\end{equation}
where $w_u,w_r\ge0$ and $w_u+w_r=1$.
We invoke the instance segmentation module when $S^{t}$ exceeds a threshold $\theta$.
To avoid using an overly outdated prior for prolonged periods, we additionally enforce a maximum skip interval:
\begin{equation}
\text{Invoke segmentation}
\quad \text{if} \quad S^{t}>\theta \ \ \text{or}\ \ \Delta t \ge N_{\max}, 
\label{eq:trigger_rule}
\end{equation}
where $\Delta t$ is the number of frames (or keyframes) since the last segmentation update.}

\paragraph{\textcolor{blue}{Coupling with temporal MP update.}}
\textcolor{blue}{When the prior is not refreshed, we update MP more conservatively to prevent overreacting to potentially stale semantic evidence:
\begin{equation}
\alpha_{\mathcal{G}}^{\prime\,t}=
\begin{cases}
\alpha_{\mathcal{G}}^{t}, & \text{if segmentation is updated at } t,\\
\rho\,\alpha_{\mathcal{G}}^{t}, & \text{otherwise}, \qquad 0<\rho<1,
\end{cases}
\label{eq:alpha_mod}
\end{equation}
and use $\alpha_{\mathcal{G}}^{\prime\,t}$ in Eq.~\eqref{eq:mp_temporal}.}

\subsection{Densification with Epipolar-Geometry Verification}\label{sec:epi_verify}
\begin{figure}[t]
\centering
\includegraphics[width=\columnwidth]{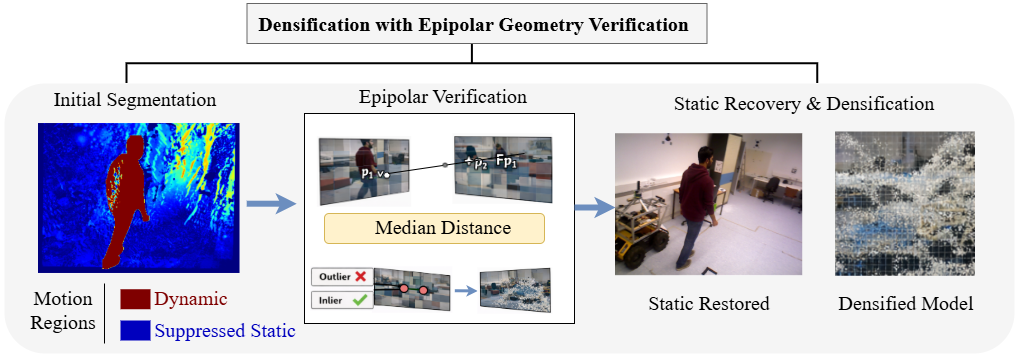}
\vspace{-2pt}
\caption{\textcolor{blue}{Epipolar-verified static recovery and densification. We first obtain a coarse MP-based dynamic mask to localize ambiguous boundary regions, then use epipolar geometry to validate static correspondences and filter dynamic outliers. The verified static regions are restored and incorporated into the densification stage.}}
\label{fig:epi}
\vspace{-4pt}
\end{figure}

\textcolor{blue}{We treat MP as a soft spatiotemporal state and use its temporally updated values to emphasize static structure in both pose refinement and Gaussian mapping. However, MP/semantic cues can be unreliable near motion boundaries and under occlusion, which may inadvertently suppress truly static structure.} To improve robustness with limited overhead, we apply epipolar-geometry verification only on \emph{ambiguous} regions (e.g., MP-suppressed Gaussians near boundaries) and densify the recovered static structure (Fig.~\ref{fig:epi}).

\paragraph{Epipolar residual.}
Given a correspondence $\tilde{\mathbf{p}}_{1}=[u_{1},v_{1},1]^{T}$ and $\tilde{\mathbf{p}}_{2}=[u_{2},v_{2},1]^{T}$ between two frames, we measure the point-to-epipolar-line distance (in pixels) as
\begin{equation}
d(\tilde{\mathbf{p}}_{2},\mathbf{F}\tilde{\mathbf{p}}_{1})
=\frac{\big|\tilde{\mathbf{p}}_{2}^{T}\mathbf{F}\tilde{\mathbf{p}}_{1}\big|}{\sqrt{a^{2}+b^{2}}},
\label{eq:epi_dist}
\end{equation}
where $\boldsymbol{\ell}_{2}=\mathbf{F}\tilde{\mathbf{p}}_{1}=[a,b,c]^T$.
We compute $\mathbf{F}$ from calibrated intrinsics and the estimated relative pose, and optionally apply RANSAC on feature matches to reject outliers when estimating/refining $\mathbf{F}$, reducing the influence of dynamic correspondences.

\paragraph{Gaussian-level decision and recovery.}
For each Gaussian $\mathcal{G}$, let $\mathcal{M}(\mathcal{G})$ be its associated verified matches.
We summarize epipolar consistency using a robust statistic
\begin{equation}
\bar d(\mathcal{G})=\mathrm{Median}\Big(\big\{d(\tilde{\mathbf{p}}_{2},\mathbf{F}\tilde{\mathbf{p}}_{1})\big\}_{(\tilde{\mathbf{p}}_{1},\tilde{\mathbf{p}}_{2})\in\mathcal{M}(\mathcal{G})}\Big).
\label{eq:agg_epi}
\end{equation}
If $\bar d(\mathcal{G})<\epsilon$, $\mathcal{G}$ is regarded as geometrically consistent with static motion.
Gaussians that are down-weighted/suppressed by MP but satisfy this criterion are restored as static, improving map continuity and reducing boundary holes.
\textcolor{blue}{The recovered static matches are then used to guide densification in the corresponding regions, following the standard 3DGS densification strategy.}

\subsection{MP-guided Rendering and Loss Optimization}\label{sec:mp_losses}
\begin{figure}[t]
\centering
\includegraphics[width=\columnwidth]{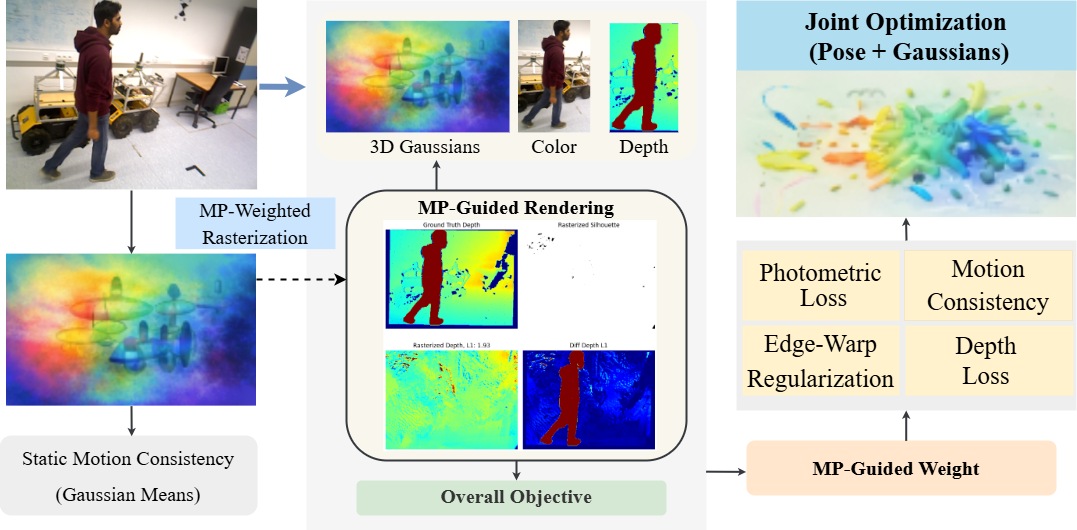}
\vspace{-2pt}
\caption{\textcolor{blue}{MP-guided rendering and joint optimization.
We convert per-Gaussian MP to a pixel-wise static confidence map to down-weight dynamic regions during rasterization.
The objective combines MP-weighted photometric/depth losses, a static motion-consistency regularizer, and an MP-guided edge-warp term to improve alignment near motion boundaries under imperfect priors.}}
\label{fig:loss}
\vspace{-4pt}
\end{figure}
Dynamic regions can dominate rendering residuals and bias pose/geometry updates (Fig.~\ref{fig:loss}).
We therefore use MP to construct \emph{soft static weights} that focus optimization on reliable static evidence while remaining smooth near uncertain boundaries.
All terms are lightweight: MP reweighting is a per-pixel scalar, and the edge-warp is computed on sparse edge pixels.

\paragraph{Pixel-wise static confidence.}
Let $M_{k}^{t}\in[0,1]$ be the MP of Gaussian $k$ at frame $t$ and $\bar M_{k}^{t}=1-M_{k}^{t}$ its static confidence.
During rasterization, Gaussian $k$ contributes to pixel $u$ with normalized weight $\omega_{k}^{t}(u)$ ($\sum_{k}\omega_{k}^{t}(u)=1$).
We define a pixel-level static weight
\begin{equation}
w^{t}(u)=\sum\nolimits_{k}\omega_{k}^{t}(u)\,\bar M_{k}^{t},
\label{eq:pixel_static_weight}
\end{equation}
which suppresses pixels dominated by dynamic Gaussians while remaining smooth around uncertain boundaries.

\paragraph{MP-weighted photometric and depth losses.}
Given rendered color/depth $C^{t}(u;\mathbf{T}_{CW}^{t})$, $D^{t}(u;\mathbf{T}_{CW}^{t})$
and aligned RGB-D references $C^{t}_{gt}(u)$, $D^{t}_{gt}(u)$, we use
\begin{equation}
L_{\mathrm{pho}}^{t}=\sum\nolimits_{u} w^{t}(u)\,\big\|C^{t}(u;\mathbf{T}_{CW}^{t})-C^{t}_{gt}(u)\big\|_{1},
\label{eq:Lpho}
\end{equation}
\begin{equation}
L_{\mathrm{depth}}^{t}=\sum\nolimits_{u} w^{t}(u)\,\big\|D^{t}(u;\mathbf{T}_{CW}^{t})-D^{t}_{gt}(u)\big\|_{1}.
\label{eq:Ldepth}
\end{equation}

\paragraph{\textcolor{blue}{MP-based motion consistency.}}
\textcolor{blue}{Instead of hard removal, we regularize Gaussians that are likely static to evolve smoothly across mapping updates.}
Let $\boldsymbol{\mu}_{k}^{t}$ be the 3D mean of Gaussian $k$ after the mapping update at frame $t$.
(Here $\boldsymbol{\mu}_{k}^{t}$ is represented in a frame-consistent coordinate system; if a camera-centric parameterization is used, the relative pose $\mathbf{T}_{t\leftarrow t-1}$ aligns the two frames.)
Using the relative pose $\mathbf{T}_{t\leftarrow t-1}\in SE(3)$, we propagate the previous mean as
\begin{equation}
\hat{\boldsymbol{\mu}}_{k}^{t}=\mathbf{T}_{t\leftarrow t-1}\,\boldsymbol{\mu}_{k}^{t-1}.
\end{equation}
We then define
\begin{equation}
L_{\mathrm{MP}}^{t}=\sum\nolimits_{k} \bar M_{k}^{t}\,\big\|\boldsymbol{\mu}_{k}^{t}-\hat{\boldsymbol{\mu}}_{k}^{t}\big\|_{2}^{2},
\label{eq:LMp}
\end{equation}
\textcolor{blue}{which mainly constrains static structure while leaving dynamic Gaussians less penalized.}

\paragraph{\textcolor{blue}{MP-guided edge-warp regularization.}}
\textcolor{blue}{To encourage local geometric consistency near motion boundaries, we add an edge-based warp term between adjacent frames $i$ and $j$.}
For an edge pixel $u\in\mathcal{E}_{i}$ in frame $i$, warp it to frame $j$ by
\begin{equation}
\mathbf{u}'=\pi\!\left(\mathbf{K}\,\mathbf{T}_{ji}\left(D_{i}(u)\,\mathbf{K}^{-1}\tilde{\mathbf{u}}\right)\right),
\label{eq:warp}
\end{equation}
where $\tilde{\mathbf{u}}=[u_x,u_y,1]^T$ and $\pi(\cdot)$ is the projection operator.
Here $D_i(u)$ denotes the aligned depth observation at pixel $u$.
We obtain $D_j(\mathbf{u}')$ and $w^{j}(\mathbf{u}')$ via bilinear sampling and define
\begin{equation}
L_{\mathrm{edge}}^{i\rightarrow j}=\sum\nolimits_{u\in\mathcal{E}_{i}}
\rho\!\left( w^{j}(\mathbf{u}')\,\big(D_{j}(\mathbf{u}')-D_{i}(u)\big)\right),
\label{eq:Ledge}
\end{equation}
\textcolor{blue}{where $\rho(\cdot)$ denotes the Huber kernel. This term focuses on reliable (static) edges and mitigates boundary artifacts under imperfect priors.}

\paragraph{Overall objective.}
We jointly optimize Gaussian parameters and camera poses by minimizing
\begin{equation}
L_{G}^{t}=\lambda_{1}L_{\mathrm{pho}}^{t}+\lambda_{2}L_{\mathrm{depth}}^{t}+\lambda_{3}L_{\mathrm{MP}}^{t}+\lambda_{4}L_{\mathrm{edge}}^{i\rightarrow j},
\label{eq:total_loss}
\end{equation}
where $\lambda_{1}$--$\lambda_{4}$ are fixed across sequences in our experiments.
Optionally, $\lambda_{3}$ can be down-weighted when the frame reliability in Sec.~\ref{sec:sod} is low to avoid over-regularization in highly uncertain cases.

\section{Experiments}
\label{sec:experiments}

\subsection{Experimental Setup}
\label{sec:exp_setup}

\textit{\textbf{a) Datasets:}} \noindent
We evaluate DAGS-SLAM on two public RGB-D benchmarks: the TUM RGB-D dataset~\cite{sturm2012benchmark} and the BONN RGB-D dataset~\cite{palazzolo2019refusion}.
TUM provides dynamic sequences from \texttt{fr3/walking/*}, featuring strong motion outliers and intermittent occlusions.
BONN exhibits stronger and more diverse dynamics (e.g., unmodeled movers, long-term occlusions, and large dynamic regions), making both tracking and reconstruction particularly challenging.

\vspace{2pt}
\textit{\textbf{b) Metrics:}} \noindent
We report absolute trajectory error (ATE, RMSE) in meters and the standard deviation (Std.) of per-frame translational errors over time after alignment.
ATE is computed following the standard TUM protocol with SE(3) alignment~\cite{sturm2012benchmark}.
We evaluate view synthesis using PSNR (dB), SSIM, and LPIPS between rendered frames and ground-truth RGB frames.
When the ground-truth model is available, we report reconstruction accuracy and completeness (cm), and the completeness ratio (percentage of points within $5$\,cm) using the BONN evaluation toolchain~\cite{palazzolo2019refusion}.
To decouple dynamic-content artifacts from static-structure quality, we also report static-region metrics by masking out dynamic pixels using a shared, external YOLO-based instance prior (same model and settings for all methods). This mask is used for evaluation only; full-frame metrics remain our primary results.
\textcolor{blue}{To analyze semantic scheduling and temporal MP update, we additionally report:
(i) \#YOLO calls per sequence,
(ii) amortized YOLO overhead (ms/frame) and its runtime fraction (YOLO\%),
and (iii) the dynamic/static label flip ratio (percentage of Gaussians whose labels change between adjacent keyframes).}

\vspace{2pt}
\textit{\textbf{c) Implementation details:}} \noindent

\textcolor{blue}{All experiments are conducted on Ubuntu 18.04 with an Intel i7 CPU and an NVIDIA RTX 3090 GPU. While we utilize a desktop GPU for standardized benchmarking, our proposed scheduler targets the reduction of semantic inference frequency—the primary bottleneck on edge devices. Thus, the reported efficiency gains are expected to scale favorably to resource-constrained platforms.}
Unless otherwise stated, we use a unified configuration across all sequences for temporal MP update and semantic-on-demand scheduling, including the bounded update rate $(\alpha_{\min},\alpha_{\max})$, the trigger threshold $\theta$, the maximum skip interval $N_{\max}$, and the update modulation factor $\rho$ (Sec.~\ref{sec:sod}).
For epipolar verification, following~\cite{cheng2022sg}, we set the epipolar distance threshold to $\epsilon{=}1$ pixel.
Camera poses are refined with Levenberg--Marquardt~\cite{ranganathan2004levenberg}.
We set loss weights to $\lambda_{1}{=}0.9$, $\lambda_{2}{=}0.1$, $\lambda_{3}{=}500$, and $\lambda_{4}{=}300$.
We keep $(\lambda_1,\lambda_2)$ consistent with the 3DGS-SLAM baseline and tune $(\lambda_3,\lambda_4)$ on a small held-out split disjoint from all sequences reported in Tables~\ref{tab:tum_ate} and \ref{tab:bonn_ate}; the final configuration is then fixed for all experiments.

\textbf{Runtime measurement.}
All runtime numbers are measured after warm-up with GPU synchronization and exclude visualization I/O, so that component timings are comparable across methods.

\vspace{2pt}
\textit{\textbf{d) Baselines:}} \noindent
We compare DAGS-SLAM with representative state-of-the-art methods, including
ORB-SLAM3~\cite{campos2021orb},
OVD-SLAM~\cite{he2023ovd},
NID-SLAM~\cite{xu2024nid},
DN-SLAM~\cite{ruan2023dn},
DDN-SLAM~\cite{li2024ddn},
Photo-SLAM~\cite{huang2024photo},
SplaTAM~\cite{keetha2024splatam},
DG-SLAM~\cite{xu2024dg},
RoDyn-SLAM~\cite{jiang2024rodyn},
Co-SLAM~\cite{wang2023co},
ESLAM~\cite{johari2023eslam},
and NICE-SLAM~\cite{zhu2022nice}.
For clarity, we group baselines into:
(i) neural/implicit-map SLAM (NID-SLAM, DN-SLAM, DDN-SLAM, Co-SLAM, ESLAM, NICE-SLAM),
(ii) Gaussian-splatting SLAM (Photo-SLAM, SplaTAM, DG-SLAM, RoDyn-SLAM),
and (iii) feature/object-aware tracking methods (ORB-SLAM3, OVD-SLAM).
We use official implementations and default parameters whenever available, adjusting only dataset-specific settings when required.

\subsection{Comparison with State-of-the-Arts on Tracking}
\label{sec:exp_tracking}

We benchmark DAGS-SLAM against representative baselines on the BONN RGB-D dataset to evaluate tracking robustness under dynamic interference.
DAGS-SLAM combines (i) MP-based dynamic/static labeling of Gaussian primitives in the back-end with (ii) a front-end epipolar-consistency check.
During correspondence selection, features in high-MP regions are adaptively down-weighted or rejected using the temporally updated MP state, which suppresses dynamic outliers and stabilizes pose estimation.
The temporal MP formulation further reduces sensitivity to transient segmentation noise and short-term occlusions, leading to more consistent pose updates.

\newcommand{\ATEtabstyle}{%
  \footnotesize
  \setlength{\tabcolsep}{3.0pt}
  \renewcommand{\arraystretch}{1.02}
}

\begin{table*}[t]
\centering
\caption{Absolute trajectory error (ATE, RMSE) on TUM RGB-D dynamic sequences (\texttt{fr3/walking/*}).}
\label{tab:tum_ate}
\ATEtabstyle

\begin{tabular*}{\textwidth}{@{\extracolsep{\fill}}
  >{\raggedright\arraybackslash}p{0.25\textwidth}
  *{6}{S[table-format=1.4] S[table-format=1.4]}
}
\toprule
\textbf{Sequence} &
\multicolumn{2}{c}{\textbf{ORB-SLAM3}} &
\multicolumn{2}{c}{\textbf{Photo-SLAM}} &
\multicolumn{2}{c}{\textbf{SplaTAM}} &
\multicolumn{2}{c}{\textbf{DDN-SLAM}} &
\multicolumn{2}{c}{\textbf{DG-SLAM}} &
\multicolumn{2}{c}{\textbf{Ours}} \\
\cmidrule(lr){2-3}\cmidrule(lr){4-5}\cmidrule(lr){6-7}\cmidrule(lr){8-9}\cmidrule(lr){10-11}\cmidrule(lr){12-13}
& {\textbf{ATE}} & {\textbf{Std.}}
& {\textbf{ATE}} & {\textbf{Std.}}
& {\textbf{ATE}} & {\textbf{Std.}}
& {\textbf{ATE}} & {\textbf{Std.}}
& {\textbf{ATE}} & {\textbf{Std.}}
& {\textbf{ATE}} & {\textbf{Std.}} \\
\midrule

\texttt{fr3/walking/xyz}
& 0.3729 & 0.1966
& 0.3054 & 0.1482
& 0.4323 & 0.1381
& 0.0148 & 0.0096
& 0.0166 & 0.0088
& \best{0.0128} & \best{0.0080} \\
\texttt{fr3/walking/xyz\_val}
& 0.6821 & 0.5125
& 0.6147 & 0.3986
& 0.8849 & 0.8096
& \best{0.0112} & 0.0071
& 0.0215 & 0.0108
& 0.0131 & \best{0.0054} \\
\texttt{fr3/walking/static}
& 0.1769 & 0.0968
& 0.1347 & 0.0118
& 0.3470 & 0.4326
& 0.0118 & 0.0060
& \best{0.0062} & 0.0101
& 0.0069 & \best{0.0032} \\
\texttt{fr3/walking/static\_val}
& 0.2784 & 0.0713
& 0.2634 & 0.0728
& 0.7368 & 0.4290
& 0.0138 & \best{0.0079}
& 0.0208 & 0.0311
& \best{0.0109} & 0.0087 \\
\texttt{fr3/walking/rpy}
& 0.6129 & 0.2696
& 0.5644 & 0.2789
& \multicolumn{2}{c}{\na}
& 0.0421 & 0.0266
& 0.0435 & \best{0.0209}
& \best{0.0318} & 0.0292 \\
\texttt{fr3/walking/rpy\_val}
& 0.3430 & 0.1982
& 0.2982 & 0.1781
& \multicolumn{2}{c}{\na}
& 0.0280 & \best{0.0159}
& 0.0389 & 0.0217
& \best{0.0266} & 0.0171 \\
\texttt{fr3/walking/half}
& 0.2321 & 0.1240
& 0.2670 & 0.1313
& 0.8900 & 0.6232
& \best{0.0214} & 0.0112
& 0.0300 & 0.0339
& 0.0229 & \best{0.0099} \\
\texttt{fr3/walking/half\_val}
& 0.4672 & 0.1906
& 0.3214 & 0.2018
& 1.3430 & 1.0567
& 0.0277 & 0.0256
& 0.0287 & 0.0381
& \best{0.0224} & \best{0.0139} \\
\midrule
\textbf{Avg.}
& 0.3957 & 0.2075
& 0.3462 & 0.1777
& 0.7723 & 0.5815
& 0.0214 & 0.0137
& 0.0258 & 0.0219
& \best{0.0184} & \best{0.0119} \\

\bottomrule
\end{tabular*}

\vspace{1pt}
{\footnotesize \textbf{Note:} Best results are highlighted in bold. All values are in meters. Averages are computed over successfully tracked sequences.}
\end{table*}

\begin{table*}[t]
\centering
\caption{Absolute trajectory error (ATE, RMSE) on the BONN RGB-D dataset.}
\label{tab:bonn_ate}
\ATEtabstyle

\begin{tabular*}{\textwidth}{@{\extracolsep{\fill}}
  >{\raggedright\arraybackslash}p{0.25\textwidth}
  *{6}{S[table-format=1.4] S[table-format=1.4]}
}
\toprule
\textbf{Sequence} &
\multicolumn{2}{c}{\textbf{ORB-SLAM3}} &
\multicolumn{2}{c}{\textbf{Photo-SLAM}} &
\multicolumn{2}{c}{\textbf{SplaTAM}} &
\multicolumn{2}{c}{\textbf{DDN-SLAM}} &
\multicolumn{2}{c}{\textbf{DG-SLAM}} &
\multicolumn{2}{c}{\textbf{Ours}} \\
\cmidrule(lr){2-3}\cmidrule(lr){4-5}\cmidrule(lr){6-7}\cmidrule(lr){8-9}\cmidrule(lr){10-11}\cmidrule(lr){12-13}
& {\textbf{ATE}} & {\textbf{Std.}}
& {\textbf{ATE}} & {\textbf{Std.}}
& {\textbf{ATE}} & {\textbf{Std.}}
& {\textbf{ATE}} & {\textbf{Std.}}
& {\textbf{ATE}} & {\textbf{Std.}}
& {\textbf{ATE}} & {\textbf{Std.}} \\
\midrule
\texttt{crowd}             & 0.4300 & 0.0210 & 0.5104 & 0.0113 & 1.8286 & 1.2384 & 0.0191 & 0.0110 & 0.0211 & 0.0180 & \best{0.0178} & \best{0.0090} \\
\texttt{crowd2}            & 0.9160 & 0.5577 & 0.9332 & 0.5789 & 3.4560 & 1.8706 & \best{0.0226} & \best{0.0109} & 0.0322 & 0.0245 & 0.0231 & 0.0127 \\
\texttt{person\_tracking}  & 0.6121 & 0.3220 & 0.3293 & 0.4518 & 0.5290 & 0.1022 & 0.0431 & \best{0.0168} & 0.0458 & 0.0340 & \best{0.0420} & 0.0179 \\
\texttt{person\_tracking2} & 0.6568 & 0.5564 & 0.3245 & 0.2681 & 0.6606 & 0.4266 & \best{0.0374} & \best{0.0128} & 0.0690 & 0.0388 & 0.0400 & 0.0219 \\
\texttt{synchronous}       & 0.7670 & 1.2382 & 0.4562 & 0.5641 & \multicolumn{2}{c}{\na} & 0.0300 & 0.0289 & 0.0491 & 0.0284 & \best{0.0280} & \best{0.0211} \\
\texttt{synchronous2}      & 1.5451 & 1.0069 & 0.6892 & 0.7763 & \multicolumn{2}{c}{\na} & 0.0186 & 0.0131 & 0.0210 & \best{0.0127} & \best{0.0178} & 0.0141 \\
\texttt{balloon}           & 0.0631 & 0.0466 & 0.0338 & 0.0176 & 2.7683 & 1.3278 & \best{0.0178} & \best{0.0090} & 0.0373 & 0.0116 & 0.0190 & 0.0103 \\
\texttt{balloon2}          & 0.4571 & 0.7818 & 0.6417 & 0.6681 & 1.2389 & 0.9862 & 0.0420 & 0.0277 & 0.0412 & 0.0340 & \best{0.0401} & \best{0.0195} \\
\midrule
\textbf{Avg.} & 0.6809 & 0.5663 & 0.4898 & 0.4170 & 1.7469 & 0.9920 & 0.0288 & 0.0163 & 0.0396 & 0.0253 & \best{0.0285} & \best{0.0158} \\
\bottomrule
\end{tabular*}

\vspace{1pt}
{\footnotesize \textbf{Note:} Best results are highlighted in bold. All values are in meters. Averages are computed over successfully tracked sequences.}
\end{table*}

On the more challenging BONN dataset, DAGS-SLAM achieves consistently lower tracking errors than most baselines (Table~\ref{tab:bonn_ate}).
We further report segmentation invocation statistics and latency breakdowns in Sec.~\ref{sec:ablation_sched}, showing that these gains are achieved with substantially reduced semantic overhead under the proposed on-demand scheduling.

\subsection{Evaluation of Gaussian Mapping}
\label{sec:exp_mapping}

In dynamic scenes, 3DGS-based SLAM may allocate Gaussians to moving objects, leading to degraded static rendering with artifacts such as ghosting and broken surfaces.
We evaluate mapping quality from three aspects: (i) qualitative rendered-view comparisons, (ii) photometric view-synthesis metrics, and (iii) geometric reconstruction metrics when ground truth is available.

\begin{figure}[t]
\centering
\setlength{\tabcolsep}{1.2pt}
\renewcommand{\arraystretch}{0}

\begin{tabular}{@{}c c c c c@{}}
  & \scriptsize\textbf{bonn\_crowd}
  & \scriptsize\textbf{bonn\_crowd2}
  & \scriptsize\textbf{fr3/wk\_xyz}
  & \scriptsize\textbf{fr3/wk\_xyz\_val} \\[-2pt]

  \adjustbox{valign=m}{\rotatebox{90}{\scriptsize\textbf{GT}}}
  & \includegraphics[width=0.23\linewidth, valign=m]{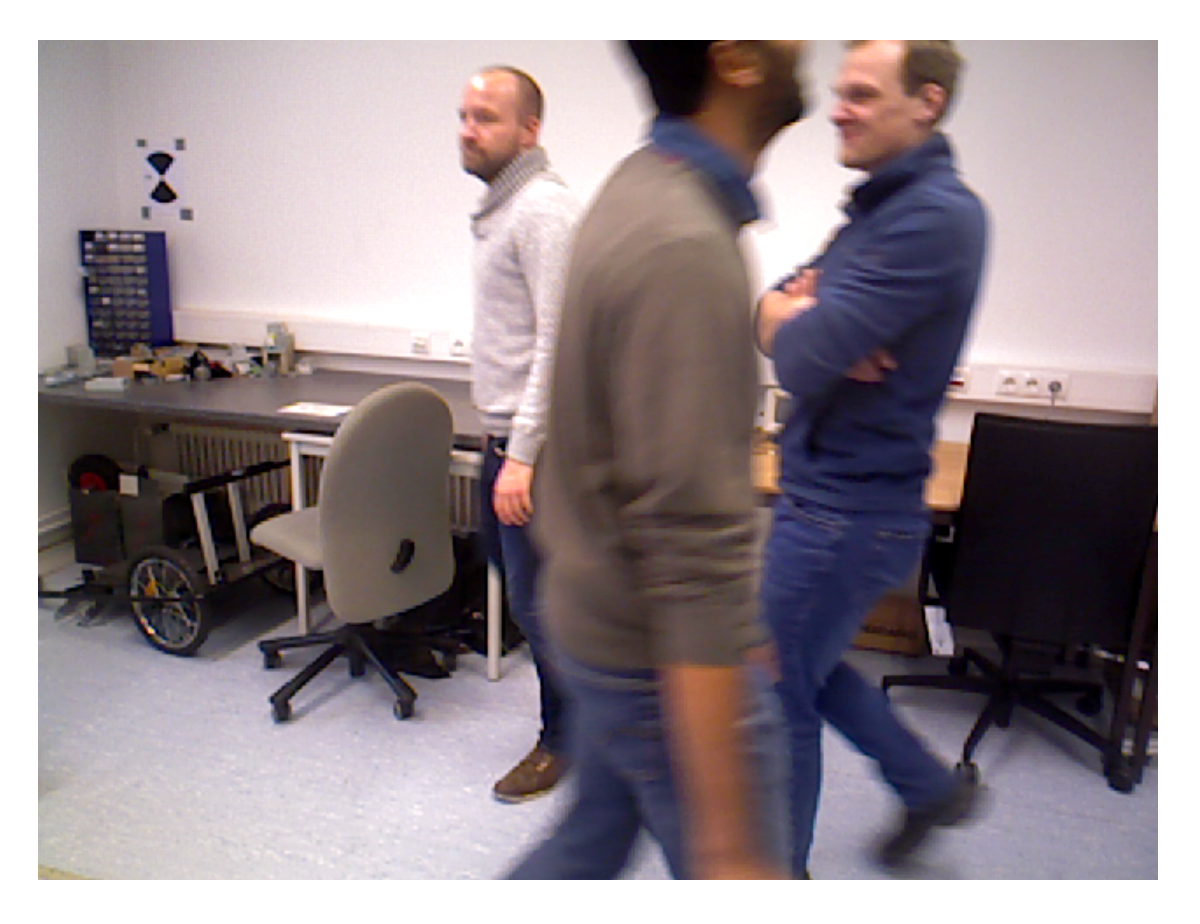}
  & \includegraphics[width=0.23\linewidth, valign=m]{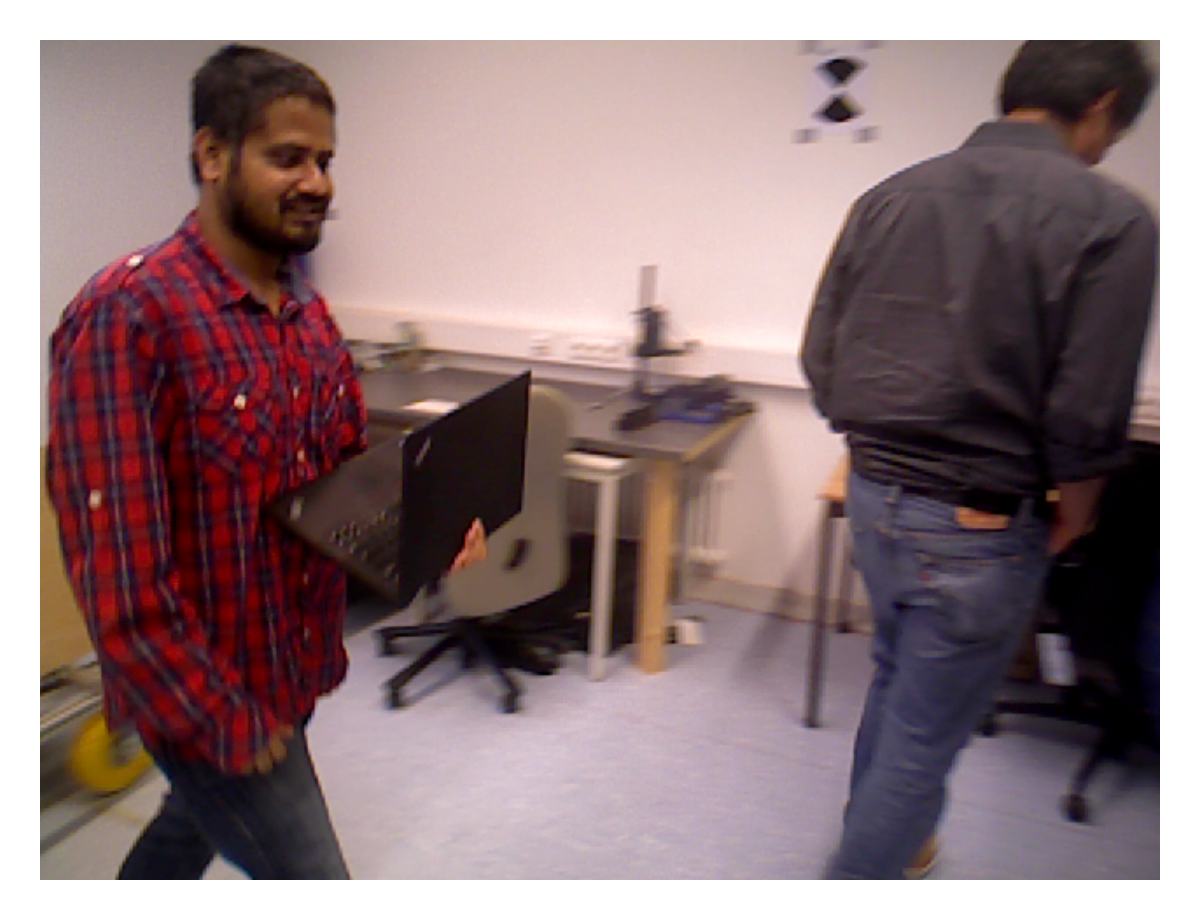}
  & \includegraphics[width=0.23\linewidth, valign=m]{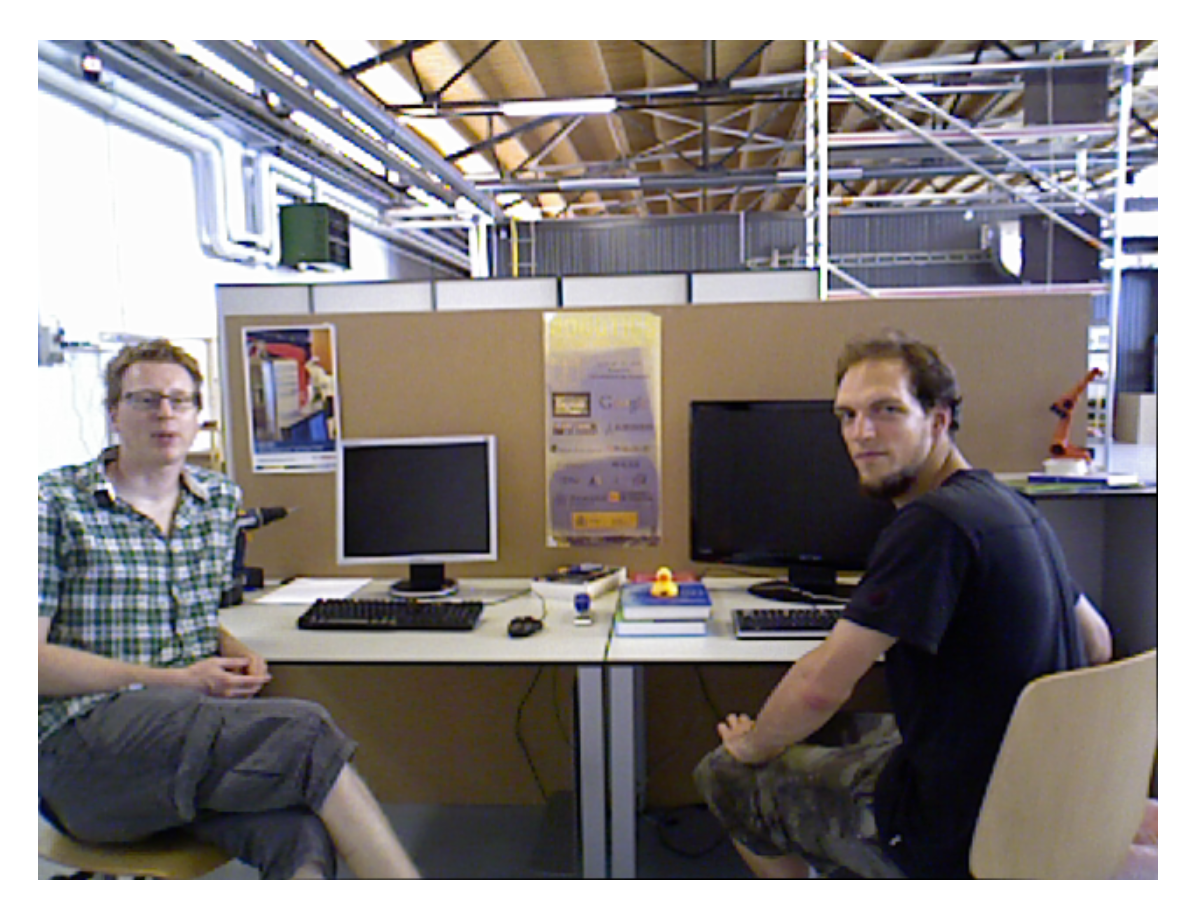}
  & \includegraphics[width=0.23\linewidth, valign=m]{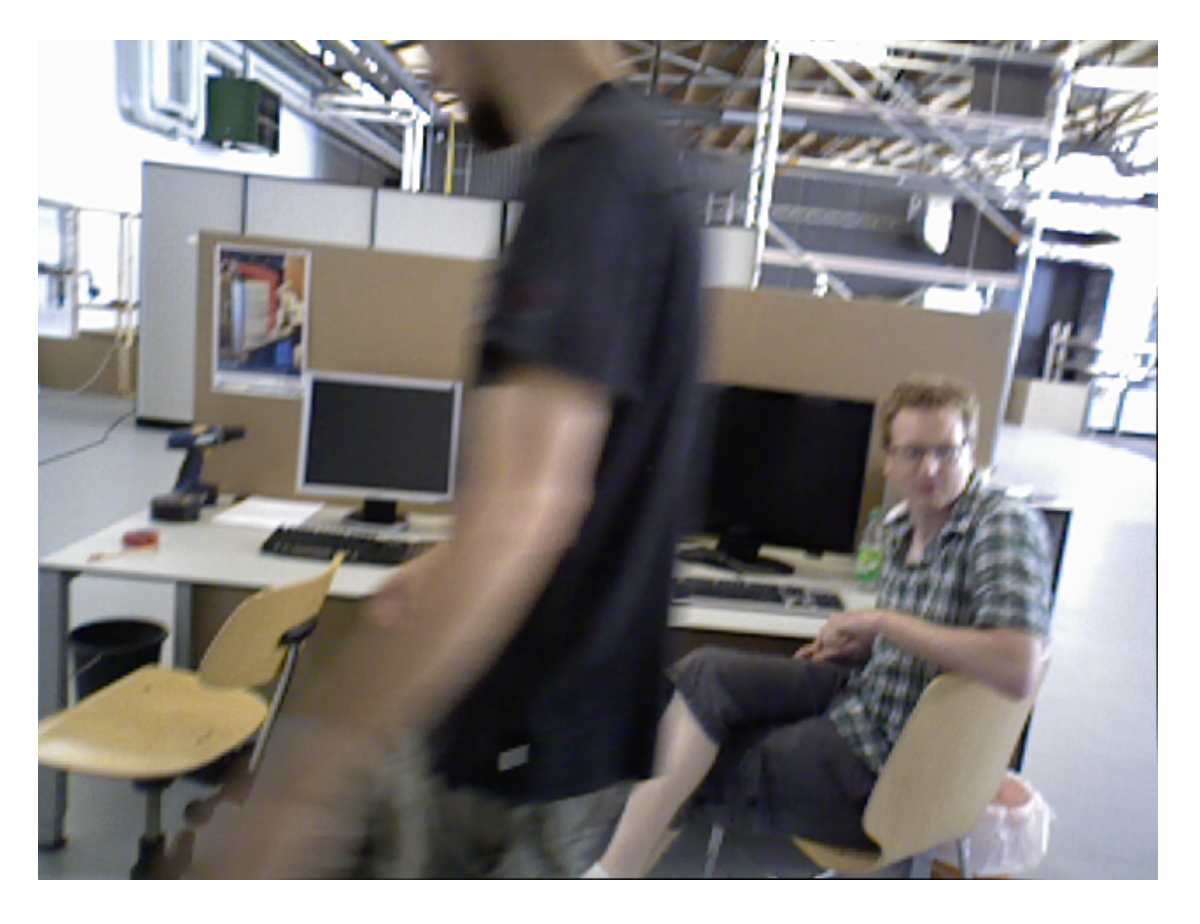} \\[-2pt]

  \adjustbox{valign=m}{\rotatebox{90}{\scriptsize\textbf{Reconstruction}}}
  & \includegraphics[width=0.23\linewidth, valign=m]{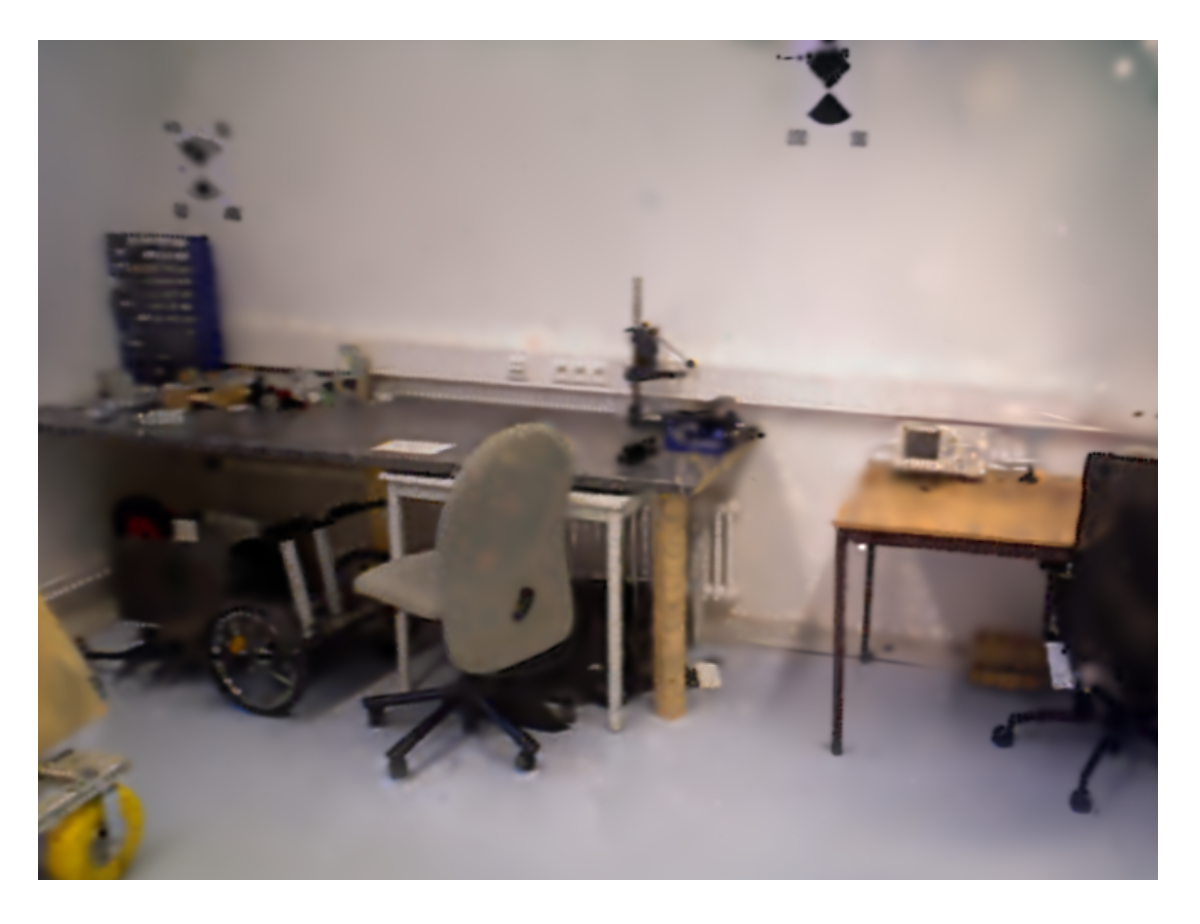}
  & \includegraphics[width=0.23\linewidth, valign=m]{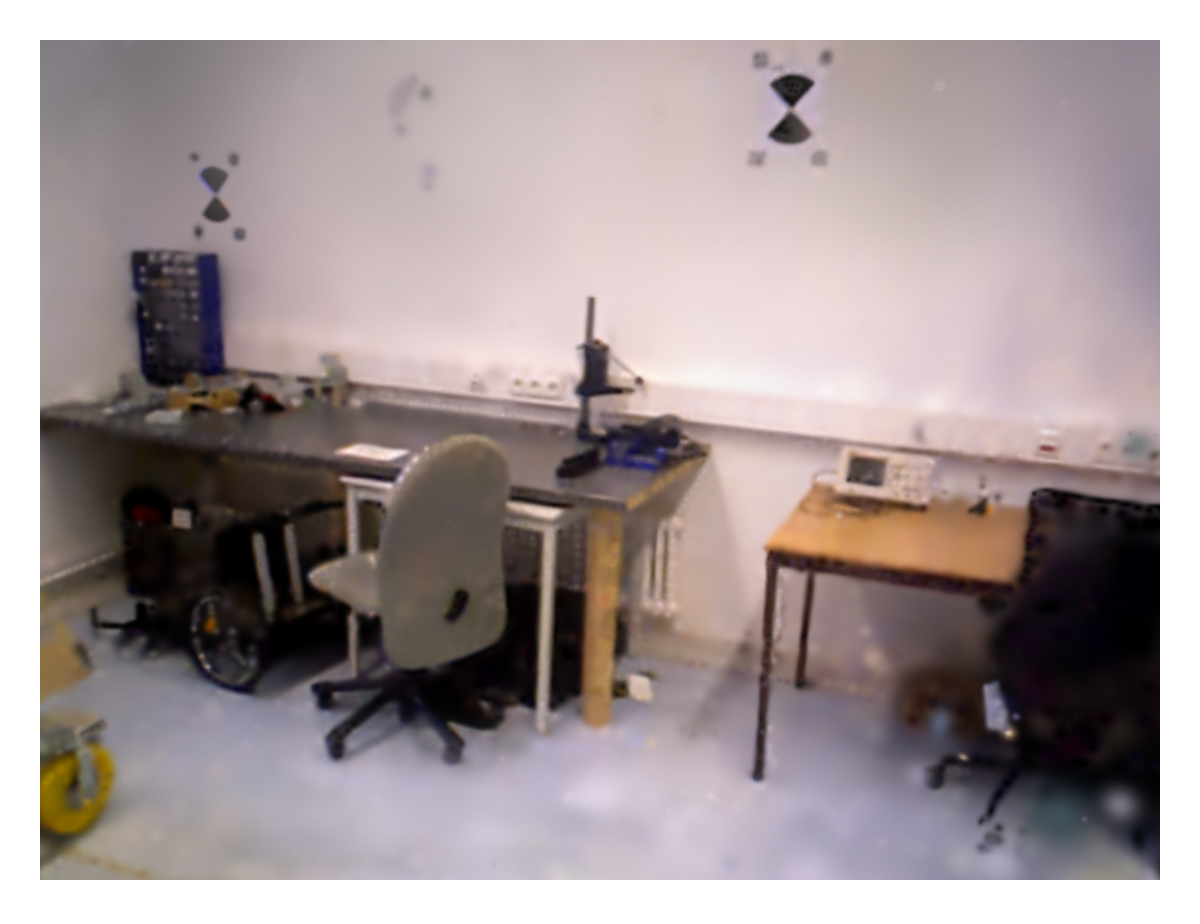}
  & \includegraphics[width=0.23\linewidth, valign=m]{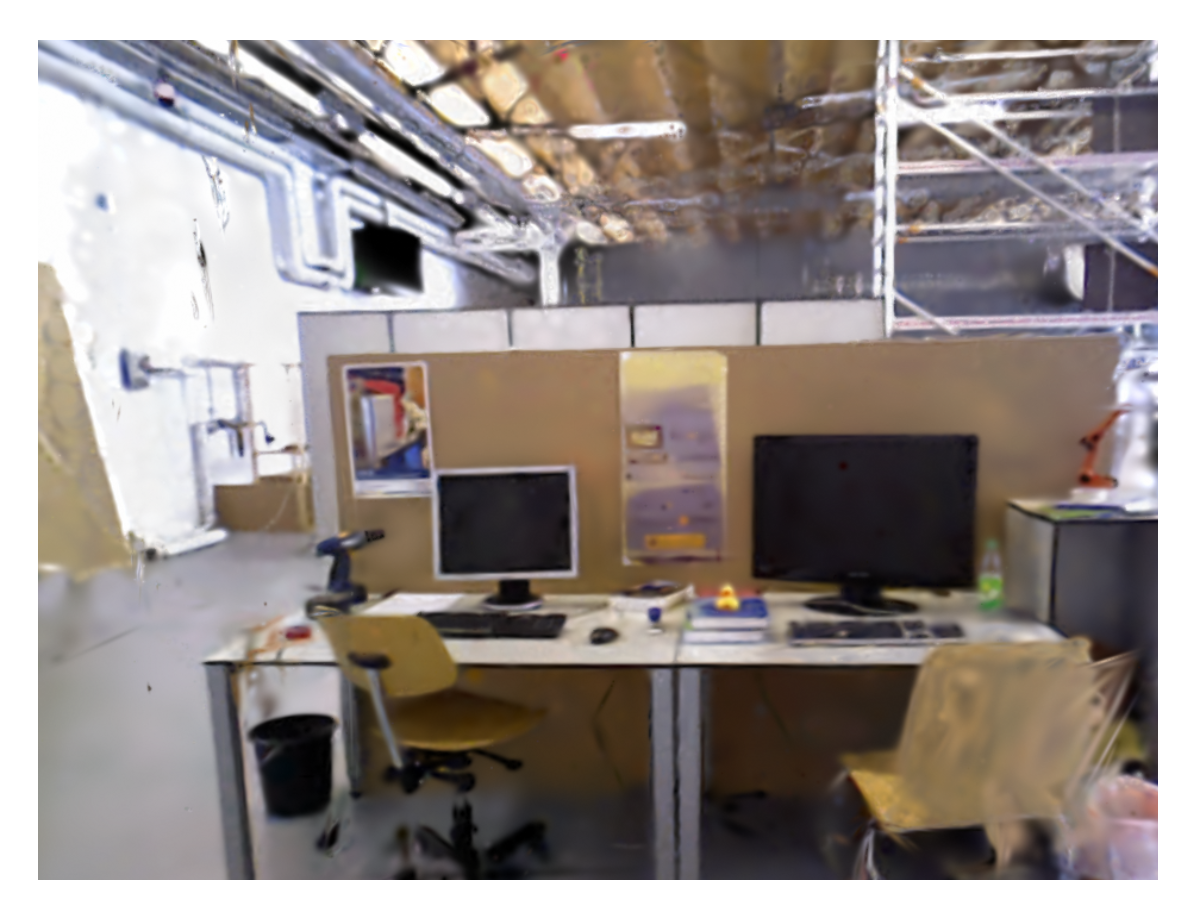}
  & \includegraphics[width=0.23\linewidth, valign=m]{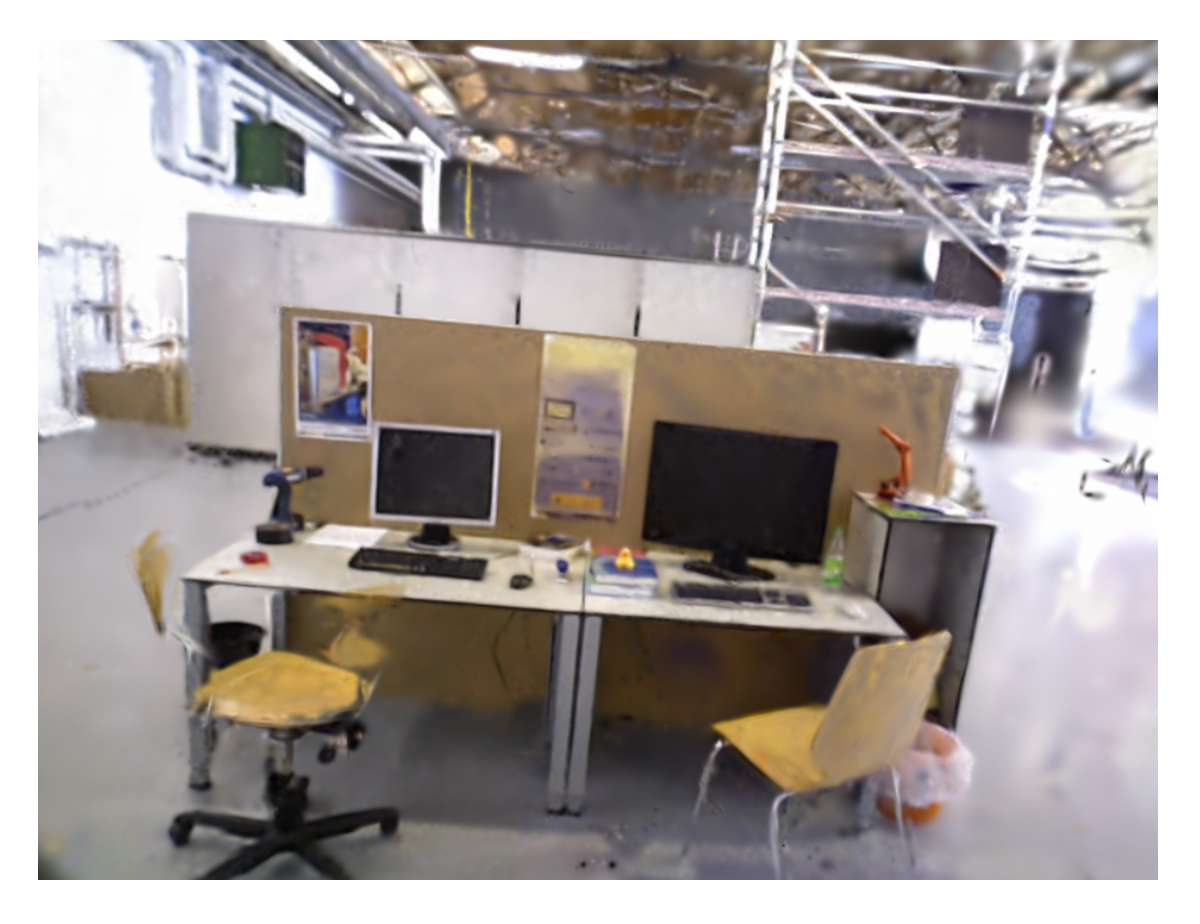} \\[-2pt]

  \adjustbox{valign=m}{\rotatebox{90}{\scriptsize\textbf{Close-up}}}
  & \includegraphics[width=0.23\linewidth, valign=m]{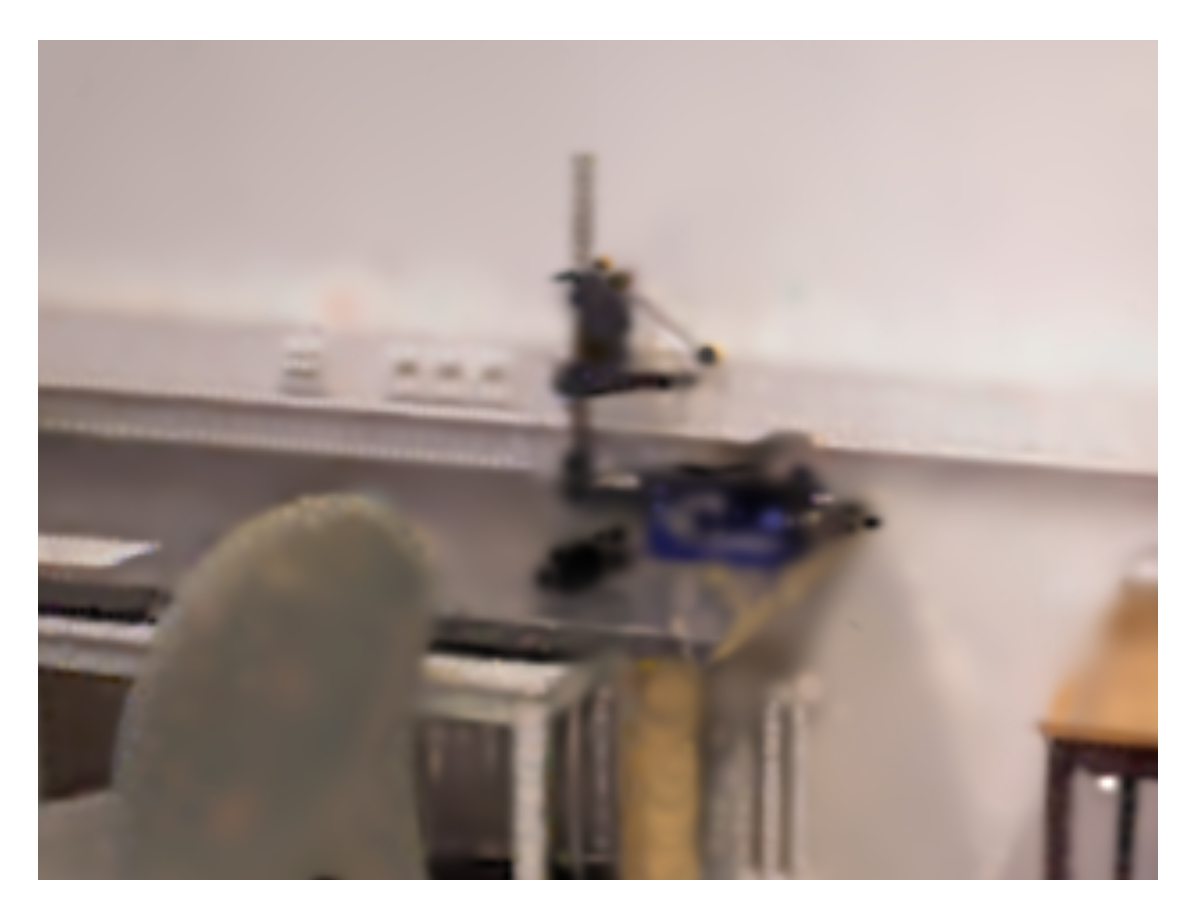}
  & \includegraphics[width=0.23\linewidth, valign=m]{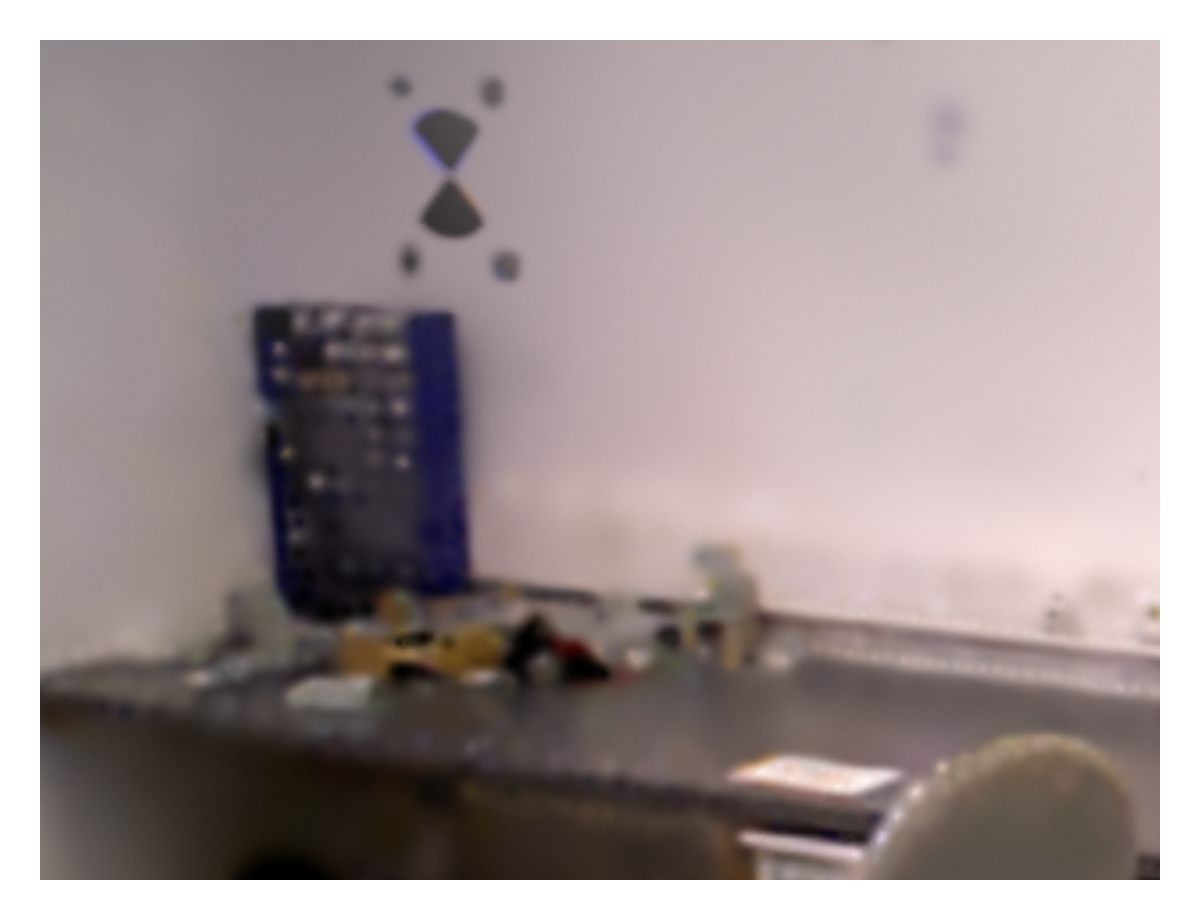}
  & \includegraphics[width=0.23\linewidth, valign=m]{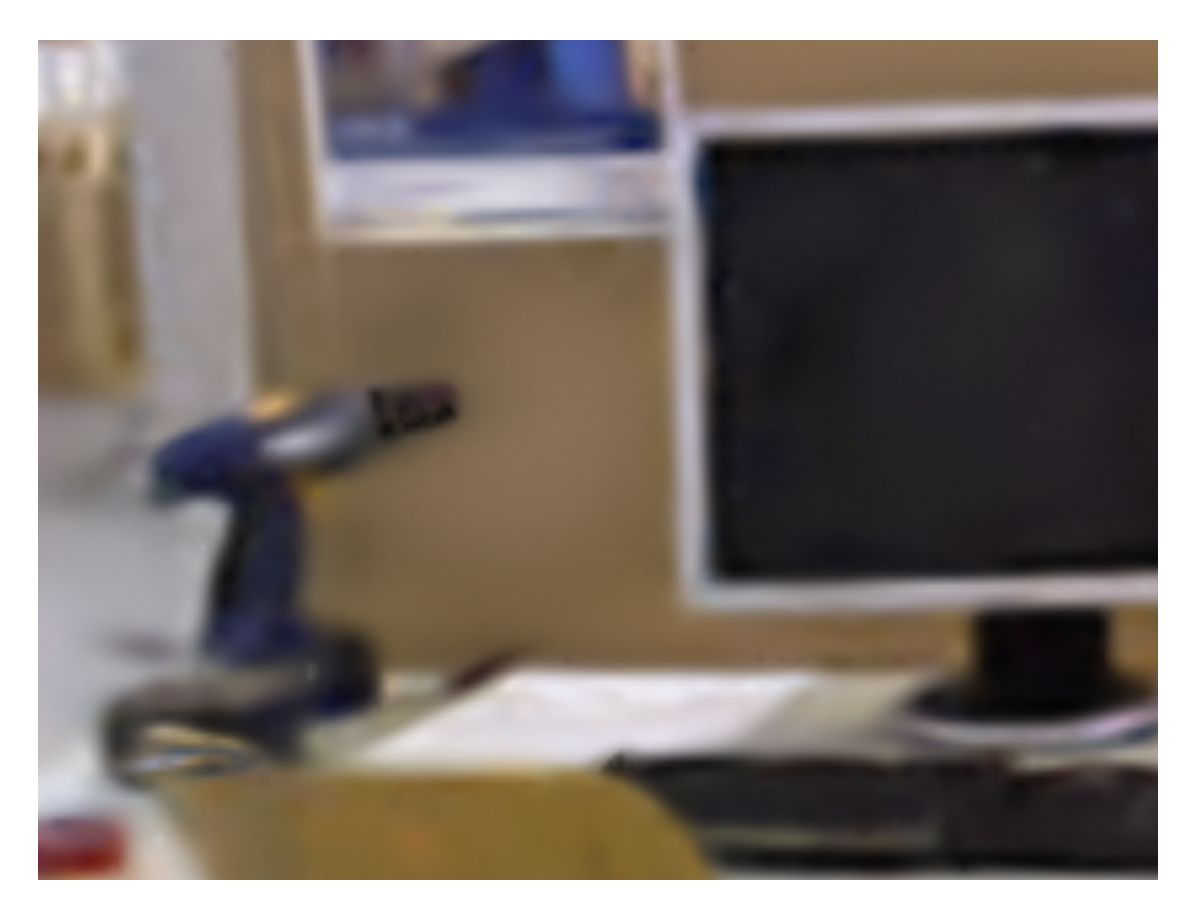}
  & \includegraphics[width=0.23\linewidth, valign=m]{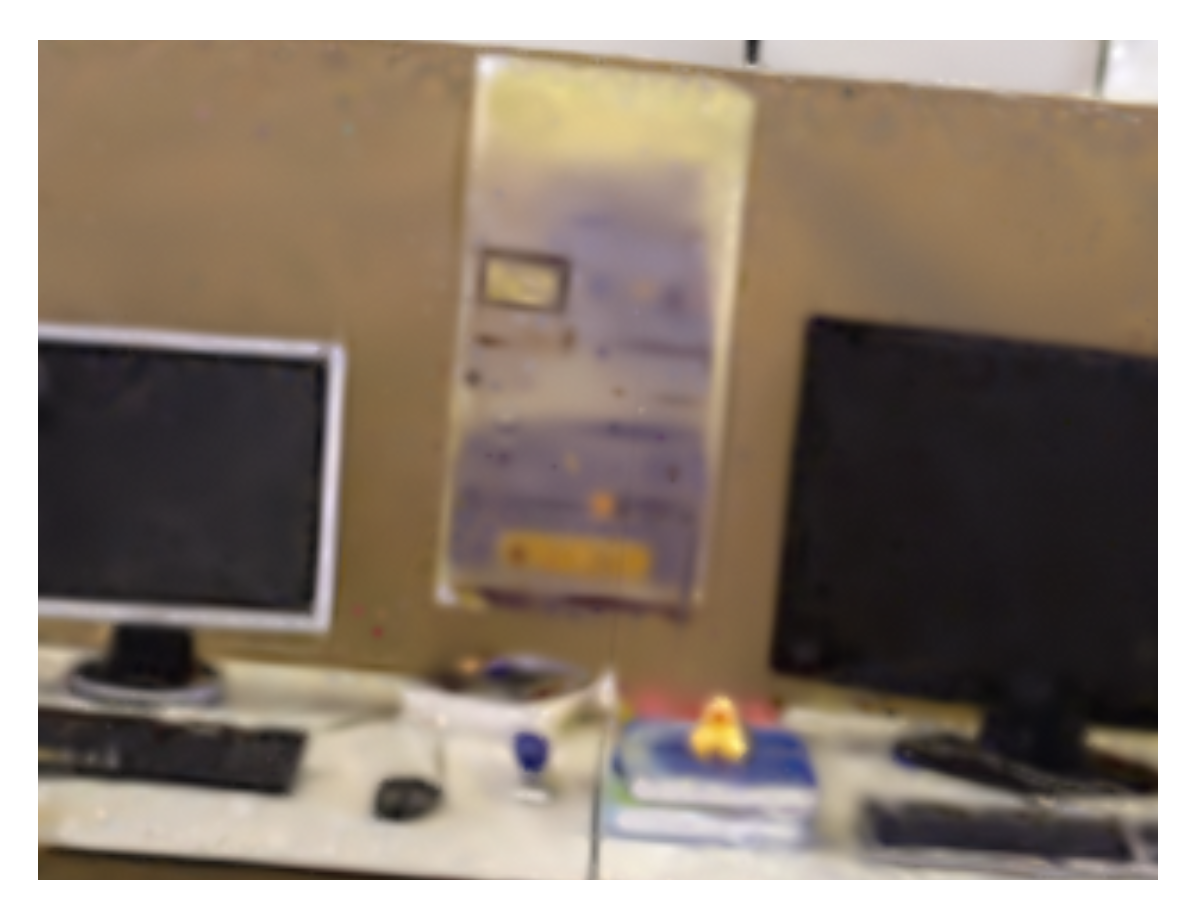}
\end{tabular}

\vspace{-3pt}
\caption{\textcolor{blue}{Qualitative visualization of DAGS-SLAM on representative dynamic sequences.}}
\label{fig:ours_gt_render_details}
\vspace{-6pt}
\end{figure}

\begin{figure}[t]
\centering
\setlength{\tabcolsep}{0pt}
\renewcommand{\arraystretch}{0}

\newlength{\gap}\setlength{\gap}{0.8pt}
\newlength{\labW}\setlength{\labW}{0.7em}
\newlength{\headH}\setlength{\headH}{0.20cm}
\newlength{\imgH}\setlength{\imgH}{1.5cm}
\newlength{\imgW}\setlength{\imgW}{\dimexpr(\linewidth-\labW-4\gap)/4\relax}

\newcommand{\CornerCell}{\begin{minipage}[c][\headH][c]{\labW}\centering\end{minipage}}
\newcommand{\ColNameTwo}[2]{%
  \begin{minipage}[c][\headH][c]{\imgW}\centering
    \scriptsize\textbf{\shortstack[c]{#1\\#2}}
  \end{minipage}}
\newcommand{\RowName}[1]{%
  \begin{minipage}[c][\imgH][c]{\labW}\centering
    \rotatebox[origin=c]{90}{\tiny\textbf{#1}}
  \end{minipage}}
\newcommand{\CellImg}[1]{%
  \begin{minipage}[c][\imgH][c]{\imgW}\centering
    \includegraphics[width=\imgW,height=\imgH]{#1}
  \end{minipage}}

\begin{tabular}{@{}c@{\hspace{\gap}}c@{\hspace{\gap}}c@{\hspace{\gap}}c@{\hspace{\gap}}c@{}}
  \CornerCell
  & \ColNameTwo{TUM}{half\_val}
  & \ColNameTwo{TUM}{xyz}
  & \ColNameTwo{BONN}{crowd}
  & \ColNameTwo{BONN}{syn2} \\[-1.5pt]

  \RowName{GT}
  & \CellImg{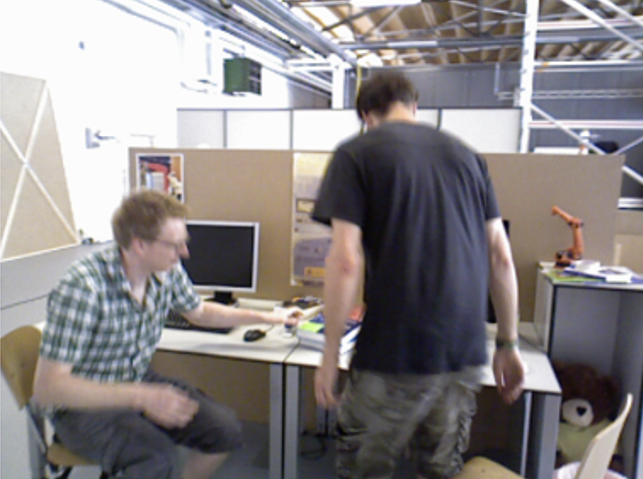}
  & \CellImg{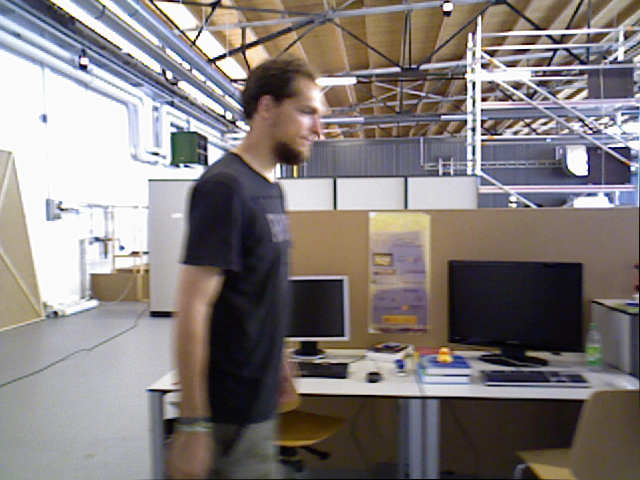}
  & \CellImg{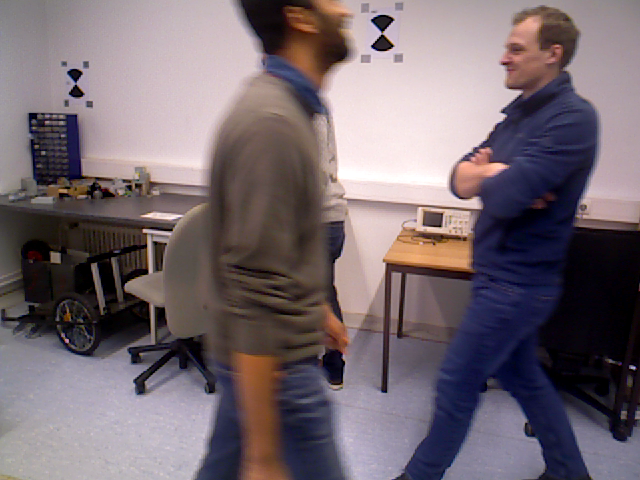}
  & \CellImg{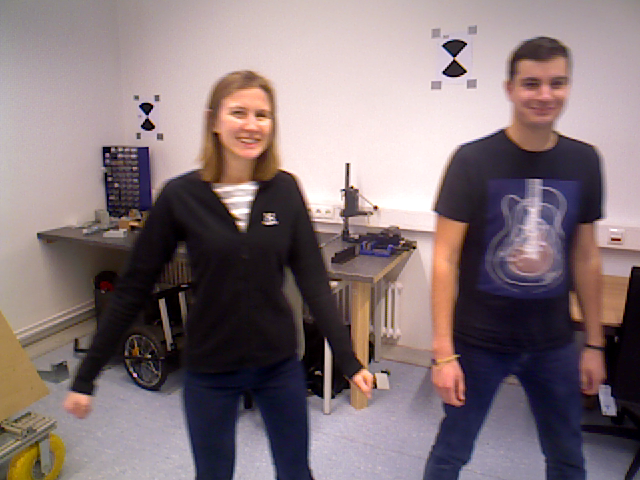} \\[-1.5pt]

  \RowName{Ours}
  & \CellImg{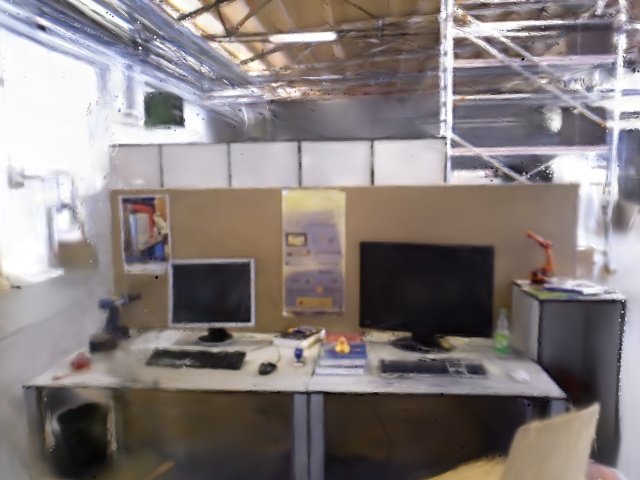}
  & \CellImg{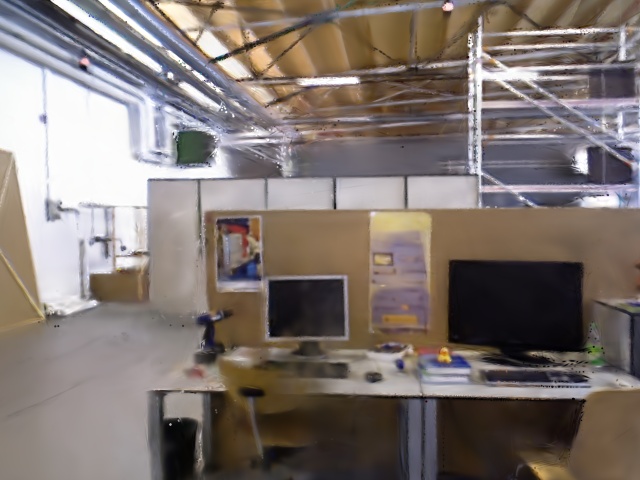}
  & \CellImg{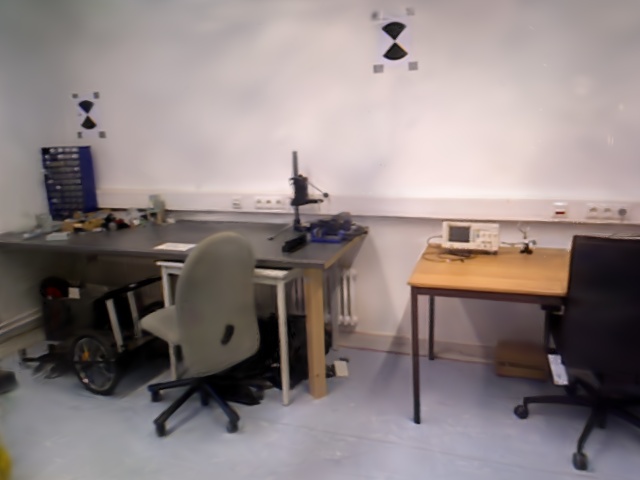}
  & \CellImg{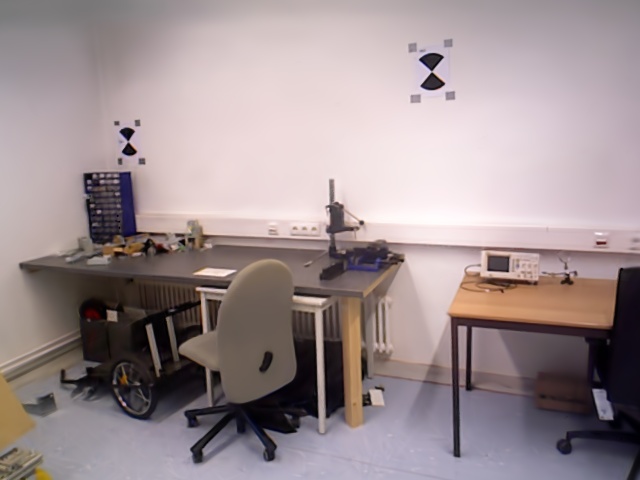} \\[-1.5pt]

  \RowName{Photo-SLAM}
  & \CellImg{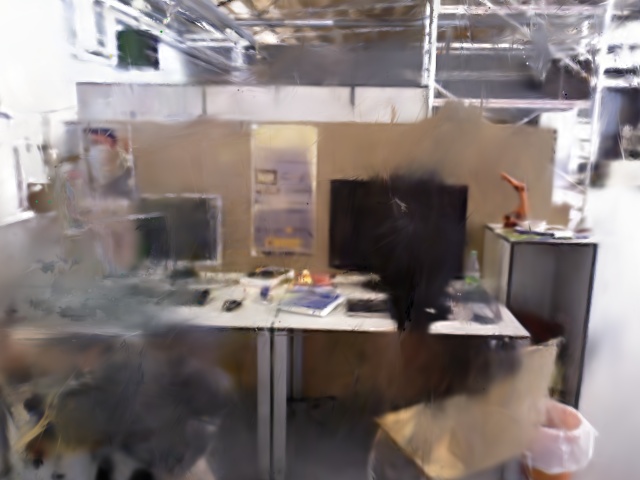}
  & \CellImg{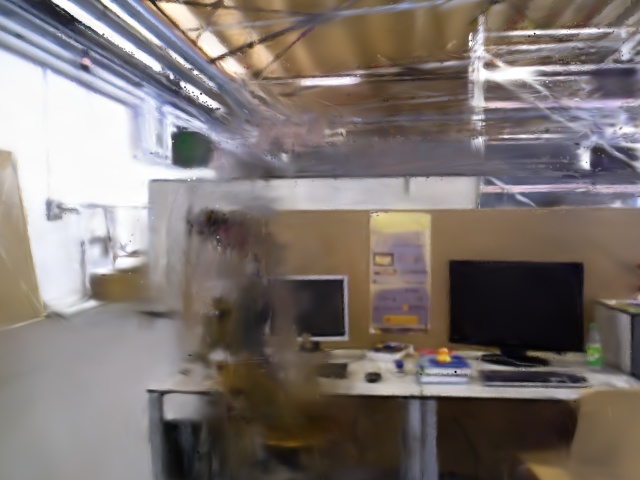}
  & \CellImg{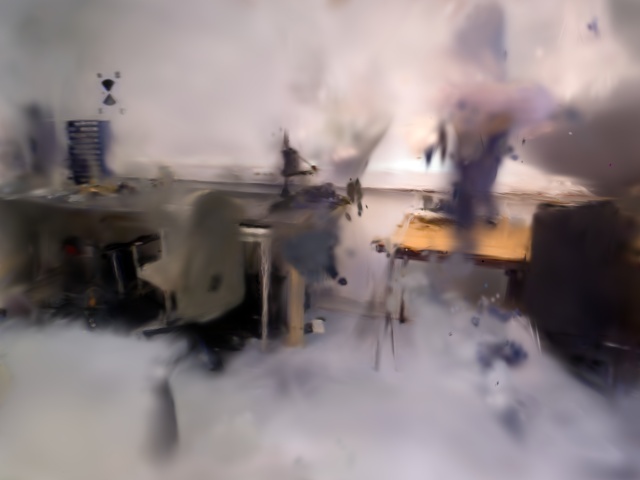}
  & \CellImg{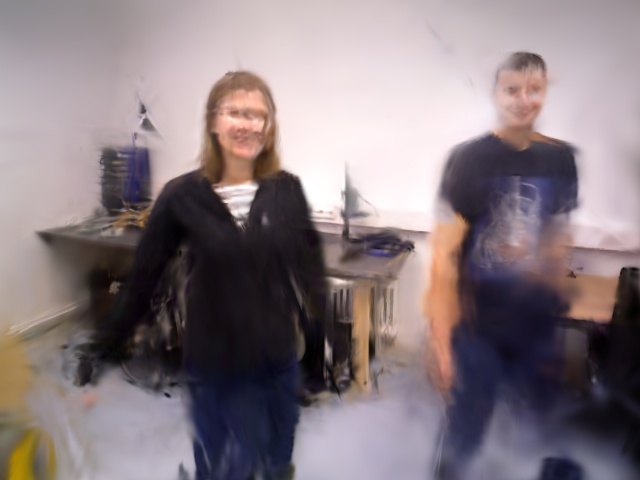} \\[-1.5pt]

  \RowName{SplaTAM}
  & \CellImg{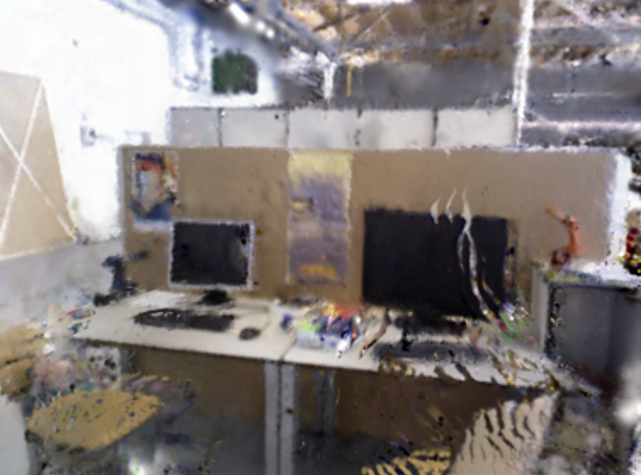}
  & \CellImg{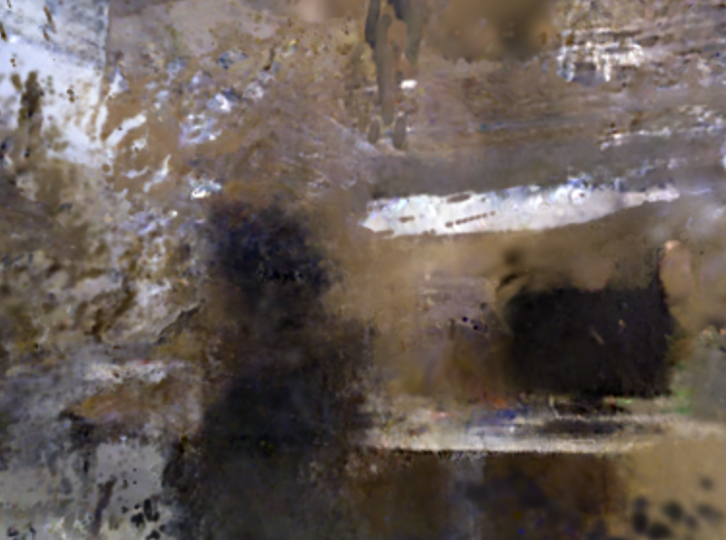}
  & \CellImg{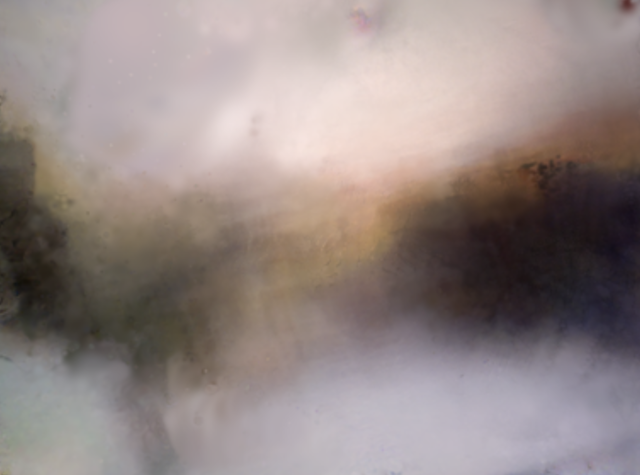}
  & \CellImg{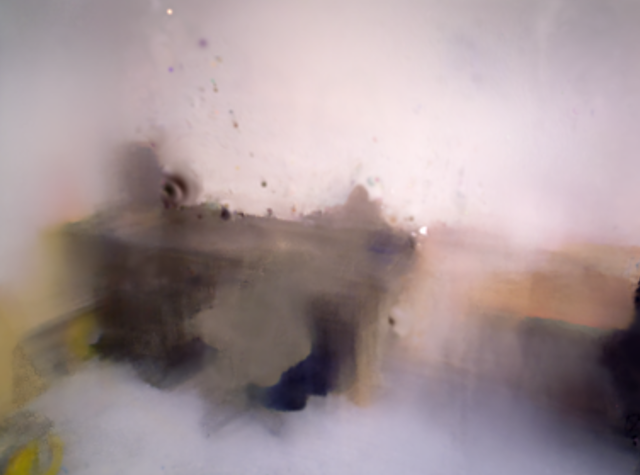} \\[-1.5pt]

  \RowName{DDN-SLAM}
  & \CellImg{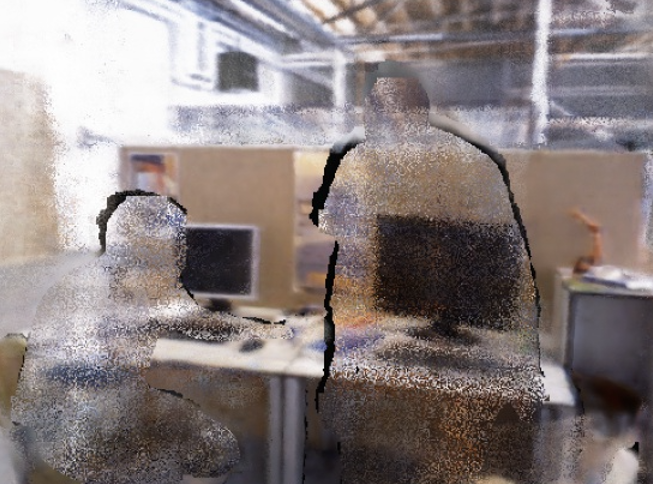}
  & \CellImg{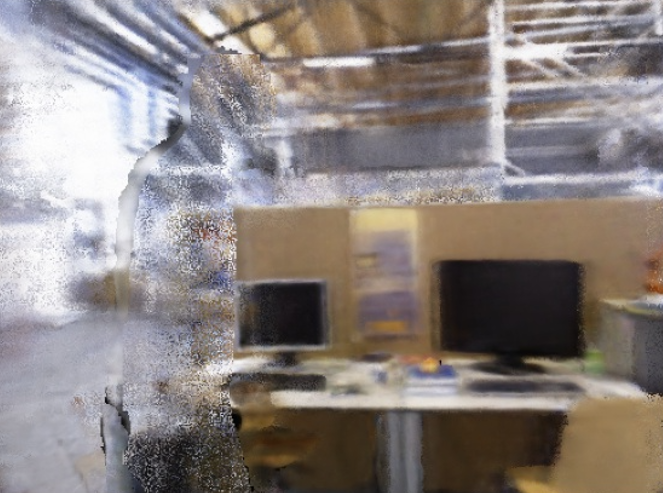}
  & \CellImg{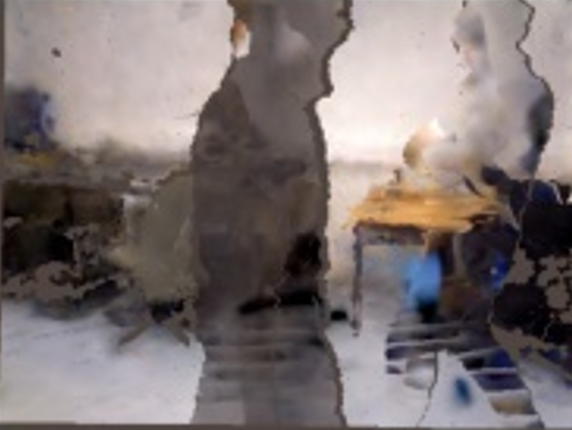}
  & \CellImg{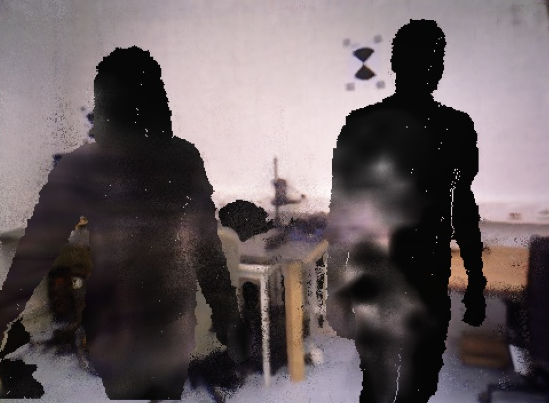} \\[-1.5pt]

  \RowName{DG-SLAM}
  & \CellImg{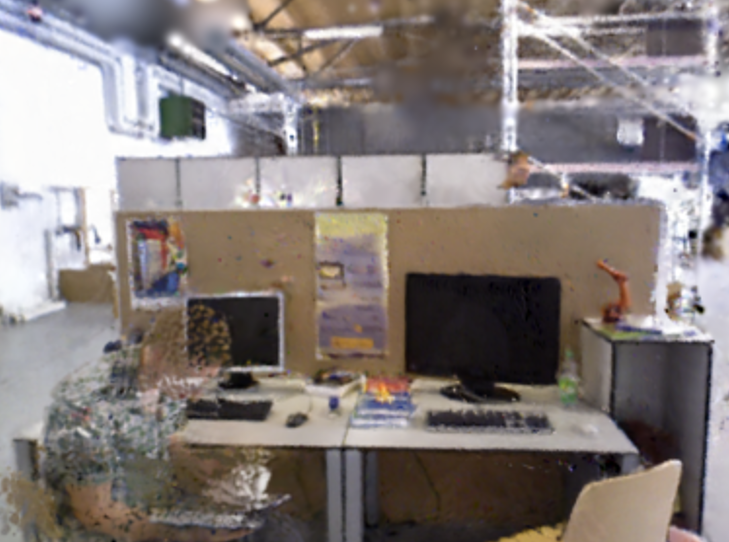}
  & \CellImg{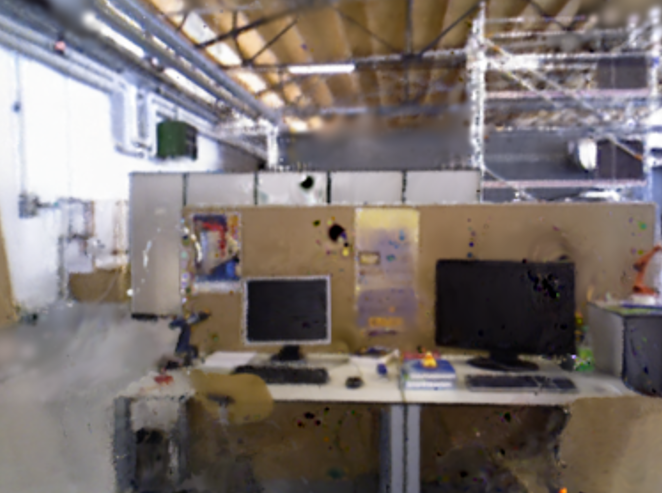}
  & \CellImg{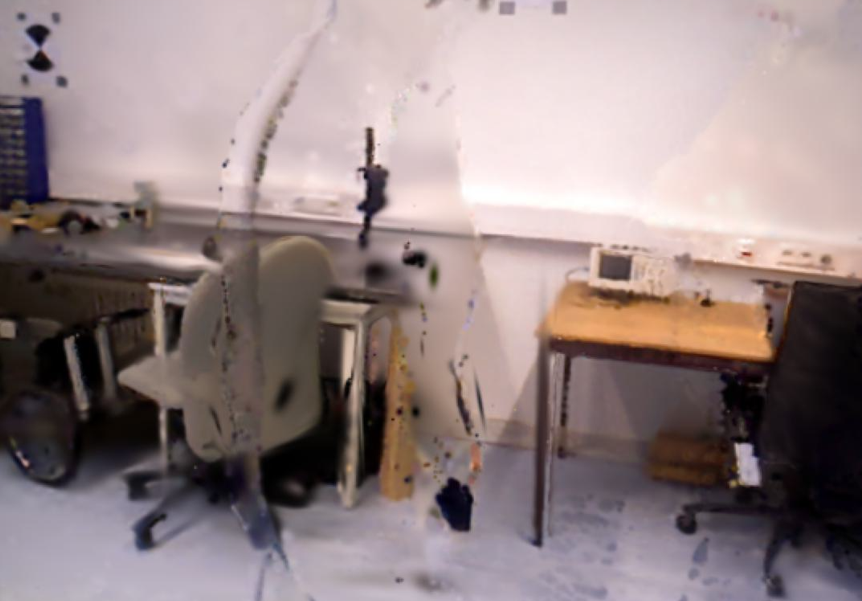}
  & \CellImg{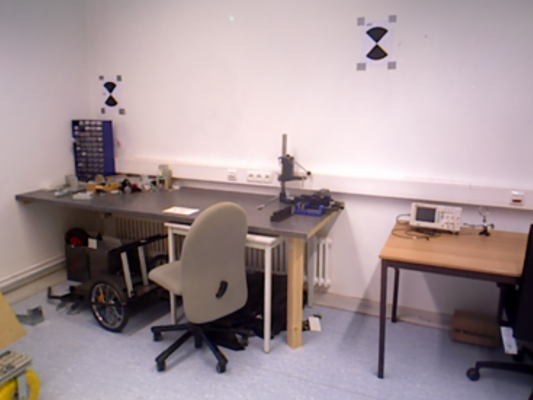} \\
\end{tabular}

\vspace{-2pt}
\caption{Qualitative comparison on representative dynamic sequences (rows: methods; columns: sequences).}
\label{fig:qual}
\vspace{-4pt}
\end{figure}

Fig.~\ref{fig:ours_gt_render_details} presents a dedicated qualitative showcase of DAGS-SLAM. The zoomed-in crops emphasize improved texture realism and boundary sharpness under occlusions and moving objects, demonstrating that our MP-guided mapping mitigates ghosting and surface breakage in dynamic scenes. As shown in Fig.~\ref{fig:qual}, DAGS-SLAM suppresses dynamic Gaussians more consistently over time, reducing ghosting and improving the continuity of static structures under occlusions and moving objects, which we attribute to MP-guided labeling with temporally smoothed MP updates.

\begin{table*}[t]
\centering
\caption{View synthesis results (mapping quality) on dynamic sequences from the BONN dataset.}
\label{tab:bonn_psnr}

\footnotesize
\setlength{\tabcolsep}{3.0pt}
\renewcommand{\arraystretch}{1.02}

\begin{tabular*}{\linewidth}{@{\extracolsep{\fill}}llcccccc}
\toprule
\textbf{Method} & \textbf{Metric} & \textbf{balloon} & \textbf{balloon2} & \textbf{person\_tracking} & \textbf{person\_tracking2} & \textbf{move\_no\_box2} & \textbf{Avg.} \\
\midrule
\multirow{3}{*}{ESLAM}
& PSNR (dB) $\uparrow$  & 15.49 & 15.40 & 14.86 & 18.79 & 16.88 & 16.28 \\
& SSIM $\uparrow$       & 0.644 & 0.662 & 0.818 & 0.647 & 0.795 & 0.713 \\
& LPIPS $\downarrow$    & 0.529 & 0.560 & 0.345 & 0.416 & 0.581 & 0.486 \\
\midrule
\multirow{3}{*}{RoDyn-SLAM}
& PSNR (dB) $\uparrow$  & 15.77 & 16.66 & 17.95 & 19.41 & 15.34 & 17.03 \\
& SSIM $\uparrow$       & 0.593 & 0.608 & 0.829 & 0.658 & 0.675 & 0.673 \\
& LPIPS $\downarrow$    & 0.396 & 0.419 & 0.326 & 0.339 & 0.500 & 0.396 \\
\midrule
\multirow{3}{*}{DG-SLAM}
& PSNR (dB) $\uparrow$  & 17.86 & 18.58 & 20.55 & 19.80 & 19.90 & 19.34 \\
& SSIM $\uparrow$       & 0.722 & \best{0.858} & 0.890 & 0.753 & \best{0.883} & 0.821 \\
& LPIPS $\downarrow$    & 0.368 & 0.433 & \best{0.202} & 0.309 & 0.458 & 0.354 \\
\midrule
\multirow{3}{*}{DAGS-SLAM}
& PSNR (dB) $\uparrow$  & \best{21.39} & \best{22.24} & \best{25.25} & \best{23.21} & \best{21.58} & \best{22.73} \\
& SSIM $\uparrow$       & \best{0.886} & 0.733 & \best{0.952} & \best{0.856} & 0.852 & \best{0.856} \\
& LPIPS $\downarrow$    & \best{0.326} & \best{0.177} & 0.352 & \best{0.166} & \best{0.240} & \best{0.252} \\
\bottomrule
\end{tabular*}

\vspace{1pt}
{\footnotesize \textbf{Note:} Best results are highlighted in bold.}
\end{table*}

Table~\ref{tab:bonn_psnr} shows that DAGS-SLAM improves view-synthesis quality over recent Gaussian-splatting baselines (e.g., DG-SLAM), producing more stable renderings under intermittent occlusions.
Because MP is used as a soft confidence, performance remains relatively stable even with coarse or intermittently refreshed instance priors under semantic-on-demand scheduling.

\begin{table*}[t]
\centering
\caption{Geometric reconstruction on dynamic scenes from the BONN dataset.}
\label{tab:bonn_geo}

\footnotesize
\setlength{\tabcolsep}{3.0pt}
\renewcommand{\arraystretch}{1.02}

\begin{tabular*}{\linewidth}{@{\extracolsep{\fill}}llcccccc}
\toprule
\textbf{Method} & \textbf{Metric} & \textbf{balloon} & \textbf{balloon2} & \textbf{person\_tracking} & \textbf{person\_tracking2} & \textbf{move\_no\_box2} & \textbf{Avg.} \\
\midrule
\multirow{3}{*}{Co-SLAM}
& Acc. [cm] $\downarrow$        & 16.70 & 31.19 & 35.41 & 54.22 & 17.12 & 30.93 \\
& Comp. [cm] $\downarrow$       & 27.29 & 40.32 & 201.40 & 118.47 & 20.47 & 81.59 \\
& Comp. Ratio [$\leq$5cm \%] $\uparrow$ & 3.11 & 3.30 & 1.98 & 2.85 & 24.91 & 7.23 \\
\midrule
\multirow{3}{*}{ESLAM}
& Acc. [cm] $\downarrow$        & 17.22 & 26.78 & 59.20 & 89.34 & 12.27 & 40.96 \\
& Comp. [cm] $\downarrow$       & 9.09 & 13.60 & 145.81 & 186.69 & 10.11 & 73.06 \\
& Comp. Ratio [$\leq$5cm \%] $\uparrow$ & 25.73 & 21.86 & 20.49 & 17.29 & 41.38 & 25.35 \\
\midrule
\multirow{3}{*}{NID-SLAM}
& Acc. [cm] $\downarrow$        & 10.58 & 14.53 & 26.58 & 25.87 & 12.69 & 18.05 \\
& Comp. [cm] $\downarrow$       & 10.71 & 30.09 & 124.92 & 107.32 & 10.19 & 56.65 \\
& Comp. Ratio [$\leq$5cm \%] $\uparrow$ & 47.71 & \best{43.97} & 11.61 & 12.59 & 40.01 & 31.18 \\
\midrule
\multirow{3}{*}{RoDyn-SLAM}
& Acc. [cm] $\downarrow$        & 10.78 & 13.33 & \best{10.45} & 13.81 & 11.56 & 11.99 \\
& Comp. [cm] $\downarrow$       & 7.21 & \best{7.90} & 27.69 & 18.87 & \best{6.79} & 13.69 \\
& Comp. Ratio [$\leq$5cm \%] $\uparrow$ & 47.60 & 40.89 & 34.07 & 32.61 & \best{45.40} & 40.11 \\
\midrule
\multirow{3}{*}{Ours}
& Acc. [cm] $\downarrow$        & \best{9.89} & \best{11.97} & 11.79 & \best{12.29} & \best{10.03} & \best{11.19} \\
& Comp. [cm] $\downarrow$       & \best{6.76} & 8.81 & \best{22.91} & \best{15.04} & 7.48 & \best{12.20} \\
& Comp. Ratio [$\leq$5cm \%] $\uparrow$ & \best{49.94} & 43.83 & \best{39.80} & \best{36.52} & 43.21 & \best{42.66} \\
\bottomrule
\end{tabular*}

\vspace{1pt}
{\footnotesize \textbf{Note:} Best results are highlighted in bold.}
\end{table*}

\begin{figure}[!t]
\centering
\setlength{\tabcolsep}{1.2pt}
\renewcommand{\arraystretch}{0.9}

\newlength{\imgWb}\setlength{\imgWb}{0.24\columnwidth}
\newlength{\imgHb}\setlength{\imgHb}{1.55cm}

\newcommand{\CellImgB}[1]{\includegraphics[width=\imgWb,height=\imgHb]{#1}}
\newcommand{\CellTxtB}[1]{\parbox[c]{\imgWb}{\centering\scriptsize\textbf{#1}}}

\begin{tabular}{@{}cccc@{}}
\CellImgB{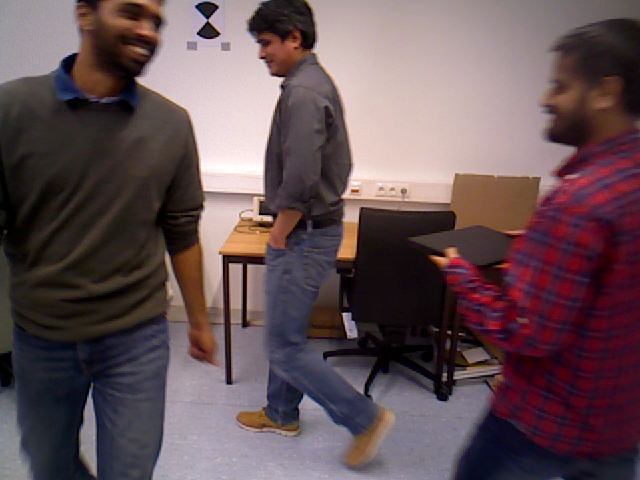} &
\CellImgB{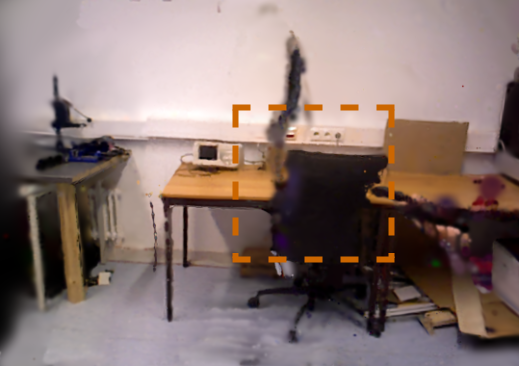} &
\CellImgB{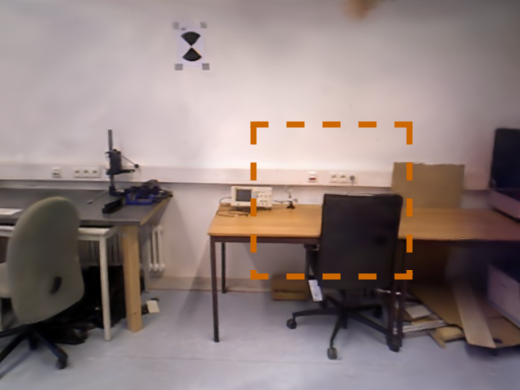} &
\CellImgB{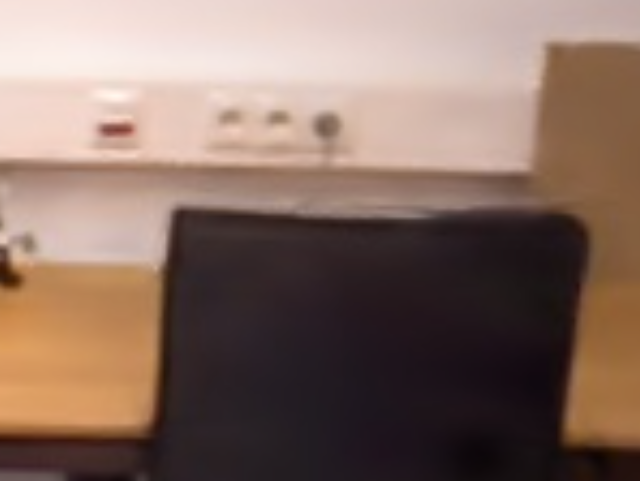} \\[-2pt]
\CellTxtB{GT} & \CellTxtB{w/o backend} & \CellTxtB{DAGS-SLAM} & \CellTxtB{Close-up} \\
\end{tabular}

\vspace{-4pt}
\caption{Rendering ablation on a representative sequence; the rightmost column shows a close-up for visual inspection.}
\label{fig:ablation_render}
\vspace{-6pt}
\end{figure}

Table~\ref{tab:bonn_geo} summarizes geometric reconstruction results on BONN, where DAGS-SLAM achieves consistently strong performance across sequences.
We further provide close-up comparisons in Fig.~\ref{fig:ablation_render} to highlight improved boundary details while avoiding over-pruning of static structures.

\subsection{Ablation Study}
\label{sec:ablation}

We ablate key components of DAGS-SLAM. Unless otherwise stated, all variants share the same implementation and hyperparameters as the full system.
For BONN, we run ablations on the subset \{\texttt{crowd}, \texttt{person\_tracking}, \texttt{move\_no\_box2}\}, where view-synthesis and geometric metrics are available and all variants are supported; results are averaged over successfully tracked sequences.
\begin{itemize}\setlength{\itemsep}{2pt}\setlength{\parskip}{0pt}
  \item \textbf{w/o MP:} removes the per-Gaussian MP state and all MP-dependent reweighting/regularization in the back-end (no MP-weighted losses, no MP regularizer, and no MP-guided correspondence filtering). Dynamic handling falls back to the raw instance prior (when available) and standard geometric checks. Since MP uncertainty is unavailable, the scheduler uses geometry-only triggering based on $R^t$.
  \item \textbf{w/o epi-verify:} disables the epipolar-geometry verification and the associated static recovery/densification near motion boundaries. All other components (MP estimation/update and MP-guided losses) remain unchanged.
  \item \textbf{w/o backend:} keeps the 3DGS mapping pipeline but removes the MP-aware back-end coupling (no association-based MP update and no MP regularizer/edge-warp). Dynamic rejection relies on the instance prior and front-end geometric checks.
\end{itemize}

\noindent\textbf{Tracking.}
Table~\ref{tab:ablation_tracking} reports ATE/Std. Removing MP substantially increases the tracking error, while disabling epipolar verification also degrades accuracy, indicating that MP-aware labeling and epipolar-based static recovery are complementary for robust pose estimation in dynamic scenes.

\begin{table}[!t]
\centering
\caption{Tracking ablation on TUM and BONN.}
\label{tab:ablation_tracking}

\footnotesize
\setlength{\tabcolsep}{3.8pt}
\renewcommand{\arraystretch}{1.05}

\begin{tabular*}{\columnwidth}{@{\extracolsep{\fill}} l l S[table-format=1.4] S[table-format=1.4] S[table-format=1.4] S[table-format=1.4]}
\toprule
\textbf{Dataset} & \textbf{Metric} & {\textbf{w/o MP}} & {\textbf{w/o epi-verify}} & {\textbf{w/o backend}} & {\textbf{Ours}} \\
\midrule
\multirow{2}{*}{BONN}
& ATE [m] $\downarrow$  & 0.5217 & 0.0451 & 0.0315 & \best{0.0212} \\
& Std. [m] $\downarrow$ & 0.3413 & 0.0276 & 0.0197 & \best{0.0135} \\
\midrule
\multirow{2}{*}{TUM}
& ATE [m] $\downarrow$  & 0.3521 & 0.0322 & 0.0230 & \best{0.0202} \\
& Std. [m] $\downarrow$ & 0.2819 & 0.0198 & 0.0226 & \best{0.0171} \\
\bottomrule
\end{tabular*}

\vspace{1pt}
{\footnotesize \textbf{Note:} Best results are highlighted in bold.}
\end{table}

\noindent\textbf{Mapping.}
Table~\ref{tab:ablation_mapping} reports view-synthesis quality on BONN (PSNR/SSIM). Disabling MP or epipolar verification degrades rendering quality, while the full system yields the best average PSNR/SSIM. We note that \texttt{crowd} SSIM is slightly higher for w/o backend, suggesting mild over-smoothing can increase SSIM without improving overall fidelity, consistent with qualitative comparisons in Fig.~\ref{fig:qual} and Fig.~\ref{fig:ablation_render}.

\begin{table}[!t]
\centering
\caption{Mapping-quality ablation on BONN.}
\label{tab:ablation_mapping}
\setlength{\tabcolsep}{5pt}
\renewcommand{\arraystretch}{1.08}
\resizebox{\linewidth}{!}{%
\begin{tabular}{l l c c c c}
\toprule
\textbf{Setting} & \textbf{Metric} & \textbf{crowd} & \textbf{person\_tracking} & \textbf{move\_no\_box2} & \textbf{Avg.} \\
\midrule
\multirow{2}{*}{w/o MP}
& PSNR (dB) $\uparrow$ & 13.63 & 18.36 & 17.32 & 16.43 \\
& SSIM $\uparrow$      & 0.612 & 0.913 & 0.730 & 0.751 \\
\midrule
\multirow{2}{*}{w/o epi-verify}
& PSNR (dB) $\uparrow$ & 14.78 & 17.91 & 19.21 & 17.30 \\
& SSIM $\uparrow$      & 0.519 & 0.816 & 0.657 & 0.664 \\
\midrule
\multirow{2}{*}{w/o backend}
& PSNR (dB) $\uparrow$ & 17.01 & 18.87 & 19.62 & 18.50 \\
& SSIM $\uparrow$      & \best{0.718} & 0.829 & 0.837 & 0.794 \\
\midrule
\multirow{2}{*}{Ours}
& PSNR (dB) $\uparrow$ & \best{22.45} & \best{24.89} & \best{20.67} & \best{22.67} \\
& SSIM $\uparrow$      & 0.708 & \best{0.923} & \best{0.885} & \best{0.839} \\
\bottomrule
\end{tabular}%
}
\vspace{2pt}
{\footnotesize \textbf{Note:} Best results are highlighted in bold.}
\end{table}

\subsubsection{Ablation I: Temporal MP and Uncertainty}
We compare (i) \textit{Instant MP}: $M_t \leftarrow \hat{M}_t$;
(ii) \textit{Temporal MP (fixed)}: $M_t \leftarrow (1-\alpha)M_{t-1} + \alpha \hat{M}_t$;
and (iii) \textit{Temporal MP (adaptive)} with uncertainty-aware $\alpha_t$ (smaller under high uncertainty, larger under consistent observations).
Table~\ref{tab:ablation_temporal_mp} reports ATE/Std. and PSNR on the same BONN subset as Table~\ref{tab:ablation_mapping}. Temporal MP improves both tracking and rendering, and the adaptive update further reduces dynamic/static label flips.

\begin{table}[!t]
\centering
\caption{Ablation on temporal MP and uncertainty-aware update.}
\label{tab:ablation_temporal_mp}
\setlength{\tabcolsep}{6pt}
\renewcommand{\arraystretch}{1.08}
\begin{tabular}{lcccc}
\toprule
\textbf{Variant} & \textbf{ATE$\downarrow$} & \textbf{Std.$\downarrow$} & \textbf{PSNR$\uparrow$} & \textbf{Flip ratio$\downarrow$} \\
\midrule
Instant MP (per-frame)   & 0.0286 & 0.0175 & 21.92 & 7.8 \\
Temporal MP (fixed)      & 0.0235 & 0.0150 & 22.34 & 5.4 \\
Temporal MP (adaptive)   & \best{0.0212} & \best{0.0135} & \best{22.67} & \best{3.6} \\
\bottomrule
\end{tabular}
\vspace{2pt}
{\footnotesize \textbf{Note:} ATE/Std. are in meters; PSNR is in dB; Flip ratio is in \%.}
\end{table}

\subsubsection{Ablation II: MP Fusion (Semantic vs. Geometric)}
We ablate MP fusion by comparing MP$_\text{ins}$ only ($K{=}0$), MP$_\text{geo}$ only ($K{=}1$), and their variance-weighted fusion (Eq.~(4)(5)).
Table~\ref{tab:ablation_mp_fusion} shows that fusion achieves the best trade-off between semantic cues and geometric robustness.

\begin{table}[!t]
\centering
\caption{Ablation on MP fusion: MP$_\text{ins}$ only vs. MP$_\text{geo}$ only vs. fusion.}
\label{tab:ablation_mp_fusion}
\setlength{\tabcolsep}{6pt}
\renewcommand{\arraystretch}{1.08}
\begin{tabular}{lcccc}
\toprule
\textbf{Variant} & \textbf{ATE$\downarrow$} & \textbf{Std.$\downarrow$} & \textbf{PSNR$\uparrow$} & \textbf{Flip$\downarrow$} \\
\midrule
MP$_\text{ins}$ only ($K{=}0$) & 0.0397 & 0.0236 & 21.35 & 8.3 \\
MP$_\text{geo}$ only ($K{=}1$) & 0.0338 & 0.0204 & 20.72 & 7.1 \\
Fusion            & \best{0.0289} & \best{0.0162} & \best{22.41} & \best{4.6} \\
\bottomrule
\end{tabular}
\vspace{2pt}

{\footnotesize \textbf{Note:} Flip denotes the dynamic/static label flip ratio (\%) of Gaussians between adjacent keyframes.}
\end{table}

\subsubsection{\textcolor{blue}{Ablation III: Semantic-on-Demand Scheduler}}
\label{sec:ablation_sched}

\textcolor{blue}{We study whether uncertainty-aware semantic scheduling reduces segmentation overhead without sacrificing accuracy.
We compare \textit{Always} (YOLO at every keyframe), \textit{Fixed-Interval} (every $N$ keyframes, $N{=}10$), and \textit{On-Demand (Ours)} (uncertainty/inconsistency trigger with max skip $N_{\max}$).
Besides ATE/Std. and PSNR/SSIM, we report segmentation calls, amortized YOLO overhead, YOLO runtime ratio, and FPS.
Table~\ref{tab:sched_tradeoff} summarizes the accuracy--efficiency trade-off and Table~\ref{tab:sched_breakdown} reports the latency breakdown.
And the on-demand scheduler achieves comparable accuracy while substantially reducing semantic overhead, resulting in higher FPS and lower total latency.}

\begin{table}[!t]
\centering
\caption{\textcolor{blue}{Accuracy--efficiency trade-off of semantic scheduling on the BONN subset.}}
\label{tab:sched_tradeoff}

\scriptsize
\setlength{\tabcolsep}{2.6pt}
\renewcommand{\arraystretch}{1.02}
\sisetup{
  detect-all,
  table-number-alignment = center,
  table-text-alignment   = center
}

\textcolor{blue}{
\begin{tabular*}{\columnwidth}{@{\extracolsep{\fill}}
l
S[table-format=1.4]
S[table-format=1.4]
S[table-format=2.2]
S[table-format=1.3]
S[table-format=3.0]
S[table-format=2.1]}
\toprule
\textbf{Sched.} &
{\textbf{ATE$\downarrow$}} &
{\textbf{Std.$\downarrow$}} &
{\textbf{PSNR$\uparrow$}} &
{\textbf{SSIM$\uparrow$}} &
{\textbf{Calls$\downarrow$}} &
{\textbf{FPS$\uparrow$}} \\
\midrule
Always           & \best{0.0210} & \best{0.0134} & \best{22.68} & \best{0.839} & 420        & 25.6 \\
Fixed ($N{=}10$) & 0.0213        & 0.0136        & 22.61        & 0.837        & 180        & 36.2 \\
On-Demand (Ours) & 0.0212        & 0.0135        & 22.67        & \best{0.839} & \best{140} & \best{40.5} \\
\bottomrule
\end{tabular*}
}

\vspace{1pt}
{\footnotesize \textbf{Note:} Calls denotes the total number of YOLO invocations summed over all sequences in this evaluation set. FPS is end-to-end; see Table~\ref{tab:sched_breakdown}.}
\end{table}

\begin{table}[!t]
\centering
\caption{\textcolor{blue}{Latency breakdown and semantic overhead under different scheduling strategies (BONN subset).}}
\label{tab:sched_breakdown}

\scriptsize
\setlength{\tabcolsep}{2.8pt}
\renewcommand{\arraystretch}{1.05}
\sisetup{
  detect-all,
  table-number-alignment = center,
  table-text-alignment   = center
}

\textcolor{blue}{
\begin{tabular*}{\columnwidth}{@{\extracolsep{\fill}}
l
S[table-format=1.1]
S[table-format=2.1]
S[table-format=1.1]
S[table-format=2.1]
S[table-format=2.1]
S[table-format=2.1]}
\toprule
\textbf{Strategy} &
\multicolumn{1}{c}{\shortstack{\textbf{Track}\\\textbf{(ms)}}} &
\multicolumn{1}{c}{\shortstack{\textbf{Map}\\\textbf{(ms)}}} &
\multicolumn{1}{c}{\shortstack{\textbf{MP}\\\textbf{(ms)}}} &
\multicolumn{1}{c}{\shortstack{\textbf{YOLO}\\\textbf{(ms)}}} &
\multicolumn{1}{c}{\shortstack{\textbf{YOLO}\\\textbf{(\%)}}} &
\multicolumn{1}{c}{\shortstack{\textbf{Total}\\\textbf{(ms)}}} \\
\midrule
Always           & 7.6 & 13.4 & 1.2 & 16.8 & 42.9 & 39.0 \\
Fixed ($N{=}10$) & 7.6 & 13.4 & 1.2 &  5.4 & 19.7 & 27.6 \\
On-Demand        & 7.6 & 13.4 & 1.2 &  2.5 & 10.1 & 24.7 \\
\bottomrule
\end{tabular*}
}

\vspace{1pt}
{\footnotesize \textbf{Note:} Per-frame average after warm-up (excluding visualization I/O). YOLO(ms) is amortized overhead.}
\end{table}

\subsection{\textcolor{blue}{Comprehensive Efficiency Analysis}}
\label{sec:efficiency}

\textcolor{blue}{We summarize overall efficiency on BONN in terms of mapping/rendering time, segmentation overhead, model size, and peak GPU memory.
DAGS-SLAM achieves a favorable efficiency profile while maintaining competitive tracking and mapping accuracy.
Note that Mapping and YOLO report component timings, while FPS is measured end-to-end; for DAGS-SLAM, the full latency breakdown (including tracking and MP update) is provided in Table~\ref{tab:sched_breakdown}.
Unless otherwise stated, \#param.\ counts the learnable parameters of the SLAM/mapping system; off-the-shelf detectors (e.g., YOLO) are excluded for clarity and fair comparison.}

\begin{table}[!t]
\centering
\caption{\textcolor{blue}{Comprehensive efficiency comparison on the BONN dataset.}}
\label{tab:efficiency_overall}

\footnotesize
\setlength{\tabcolsep}{3.2pt}
\renewcommand{\arraystretch}{1.03}

\sisetup{
  detect-all,
  table-number-alignment = center,
  table-text-alignment   = center
}

\textcolor{blue}{
\begin{tabular*}{\columnwidth}{@{\extracolsep{\fill}} l
  S[table-format=3.1]
  S[table-format=2.1]
  S[table-format=3.0]
  S[table-format=2.1]
  S[table-format=2.1]
  S[table-format=2.1]}
\toprule
\textbf{Method}
& \multicolumn{1}{c}{\textbf{\shortstack[c]{Map.\\(ms)$\downarrow$}}}
& \multicolumn{1}{c}{\textbf{\shortstack[c]{YOLO\\(ms)$\downarrow$}}}
& \multicolumn{1}{c}{\textbf{\shortstack[c]{\#YOLO\\$\downarrow$}}}
& \multicolumn{1}{c}{\textbf{\shortstack[c]{FPS$\uparrow$}}}
& \multicolumn{1}{c}{\textbf{\shortstack[c]{Params\\(M)$\downarrow$}}}
& \multicolumn{1}{c}{\textbf{\shortstack[c]{GPU\\(GB)$\downarrow$}}} \\
\midrule
NICE-SLAM & 291.8 & \multicolumn{1}{S}{\na} & \multicolumn{1}{S}{\na} & 1.2  & 11.9 & 10.9 \\
DN-SLAM   & 61.9  & \multicolumn{1}{S}{\na} & \multicolumn{1}{S}{\na} & 4.2  & 30.1 & \best{5.8} \\
DDN-SLAM  & 50.7  & \multicolumn{1}{S}{\na} & \multicolumn{1}{S}{\na} & 20.1 & 19.6 & 6.2 \\
\textbf{Ours} & \best{13.4} & \best{2.5} & \best{140} & \best{40.5} & \best{10.2} & 6.4 \\
\bottomrule
\end{tabular*}
}

\vspace{1pt}
{\footnotesize \textbf{Note:}
Best results are highlighted in bold. Map.(ms) and YOLO(ms) report per-frame component timings (excluding tracking and MP update).}
\end{table}

\subsection{\textcolor{blue}{Robustness Tests}}
\label{sec:robustness}

\textcolor{blue}{We evaluate robustness under three degradations: corrupted instance priors, strong occlusion, and fast motion.
Unless otherwise stated, all settings follow Sec.~\ref{sec:exp_setup}.}

\subsubsection{\textcolor{blue}{Robustness to Corrupted Instance Priors}}
\textcolor{blue}{To emulate imperfect priors, we corrupt YOLO masks via (1) dilation/erosion, (2) random dropout, and (3) temporal delay.
We apply the same corruption protocol to mask-dependent baselines; due to space constraints, we report DG-SLAM as a representative mask-based baseline here.
Mild and Severe indicate increasing perturbation strength. Table~\ref{tab:robust_mask} shows that DAGS-SLAM degrades gracefully as corruption increases, consistent with temporal MP smoothing and uncertainty-aware updates/scheduling.}

\begin{table}[!t]
\centering
\caption{\textcolor{blue}{Robustness to corrupted instance priors on BONN dataset.}}
\label{tab:robust_mask}
\setlength{\tabcolsep}{4pt}
\renewcommand{\arraystretch}{1.05}
\textcolor{blue}{
\resizebox{\linewidth}{!}{
\begin{tabular}{lcccccc}
\toprule
\textbf{Method} & \textbf{Corruption}
& \textbf{ATE$\downarrow$} & \textbf{Std.$\downarrow$}
& \textbf{PSNR$\uparrow$} & \textbf{SSIM$\uparrow$} & \textbf{LPIPS$\downarrow$} \\
\midrule
DG-SLAM & None   & 0.0392 & 0.0238 & 18.10 & 0.767 & 0.378 \\
DG-SLAM & Mild   & 0.0615 & 0.0369 & 16.92 & 0.728 & 0.441 \\
DG-SLAM & Severe & 0.0964 & 0.0578 & 14.83 & 0.661 & 0.523 \\
\midrule
DAGS-SLAM & None   & \best{0.0286} & \best{0.0157} & \best{22.73} & \best{0.855} & \best{0.252} \\
DAGS-SLAM & Mild   & \best{0.0314} & \best{0.0179} & \best{22.10} & \best{0.846} & \best{0.268} \\
DAGS-SLAM & Severe & \best{0.0398} & \best{0.0241} & \best{20.82} & \best{0.822} & \best{0.310} \\
\bottomrule
\end{tabular}}
}
\vspace{2pt}
{\footnotesize \textbf{Note:} Best results are highlighted in bold. ``None'' uses original masks; ``Mild/Severe'' indicate increasing corruption.}
\end{table}

\subsubsection{\textcolor{blue}{Occlusion Stress Test}}
\textcolor{blue}{We evaluate occlusion-heavy segments where dynamic regions occupy a large fraction of the view (e.g., mask-area or dynamic-Gaussian ratio), and report tracking/rendering metrics on these segments.
DAGS-SLAM shows fewer drift bursts under long occlusions, benefiting from temporal MP smoothing and epipolar verification near dynamic boundaries.}

\subsubsection{\textcolor{blue}{Fast Motion and Motion Blur}}
\textcolor{blue}{We stress-test fast motion using rapid camera/object motions and temporal subsampling to increase inter-frame displacement.
We report ATE/Std. and view-synthesis metrics under different skip factors.
DAGS-SLAM remains competitive, as MP-based penalties enforce static-structure consistency when per-frame priors become unreliable.}

\section{Conclusion}
We presented DAGS-SLAM, a dynamic-robust Gaussian-splatting SLAM system designed for mobile and edge robots/AR devices that require real-time localization and dense reconstruction under tight compute and energy budgets. DAGS-SLAM tightly couples tracking and Gaussian mapping by maintaining a temporally consistent motion-probability (MP) state with uncertainty-aware updates, and by integrating MP fusion with front-end epipolar validation to suppress dynamic interference for stable pose estimation. \textcolor{blue}{To make dynamic handling affordable on resource-constrained platforms, we further introduce an uncertainty-aware semantic-on-demand scheduler that invokes instance segmentation only when needed, substantially reducing semantic overhead while preserving accuracy.} Experiments on TUM RGB-D and BONN demonstrate improved tracking robustness and higher-quality static reconstructions with fewer dynamic artifacts, \textcolor{blue}{together with a favorable accuracy--efficiency trade-off enabled by reduced semantic invocations.}
Future work will focus on further latency/energy optimization and broader validation on real mobile platforms across diverse dynamic scenarios.

\bibliographystyle{IEEEtran}
\bibliography{references}

\end{document}